\documentclass[lettersize,journal]{IEEEtran}
\usepackage{amsmath,amsfonts}
\usepackage{algorithmic}
\usepackage{algorithm}
\usepackage{array}
\usepackage[caption=false,font=normalsize,labelfont=sf,textfont=sf]{subfig}
\usepackage{textcomp}
\usepackage{stfloats}
\usepackage{url}
\usepackage{verbatim}
\usepackage{graphicx}
\usepackage{cite}
\usepackage{tabularx,booktabs}
\usepackage{graphics}
\usepackage{graphicx}
\usepackage{wrapfig}
\usepackage{multicol}
\usepackage{caption}
\usepackage{arydshln}
\usepackage{arabtex}
\usepackage{utf8}
\setcode{utf8}

\makeatletter

\def\XLSR{{$\tt XLS\text{-}R$}}

\def\muxlsr{{$\tt SAMU\text{-}XLSR$}}
\def\labse{{$\tt LaBSE$}}
\def\asr{{$\tt ASR$}}

\begin{document}

\title{SAMU-XLSR: Semantically-Aligned Multimodal Utterance-level Cross-Lingual Speech Representation}

\author{Sameer Khurana$^1$, Antoine Laurent$^2$, James Glass$^1$\\
        $^1$MIT Computer Science and Artificial Intelligence Laboratory, Cambridge, MA, USA\\
        $^2$LIUM - Le Mans University, France\\
        \{skhurana, glass\}@mit.edu
\thanks{}
\thanks{Preprint. Under Review.}}

\markboth{}%
{Shell \MakeLowercase{\textit{et al.}}: A Sample Article Using IEEEtran.cls for IEEE Journals}


\maketitle

\begin{abstract}
We propose the \muxlsr{}: \underline{S}emantically-\underline{A}ligned \underline{M}ultimodal \underline{U}tterance-level \underline{Cross}-\underline{L}ingual \underline{S}peech \underline{R}epresentation learning framework. Unlike previous works on speech representation learning, which learns multilingual contextual speech embedding at the resolution of an acoustic frame (10-20ms), this work focuses on learning multimodal (speech-text) multilingual speech embedding at the resolution of a sentence (5-10s) such that the embedding vector space is semantically aligned across different languages. We combine state-of-the-art multilingual acoustic frame-level speech representation learning model \XLSR{} with the Language Agnostic BERT Sentence Embedding (\labse{}) model to create an utterance-level multimodal multilingual speech encoder \muxlsr{}. Although we train \muxlsr{} with only multilingual transcribed speech data, cross-lingual speech-text and speech-speech associations emerge in its learned representation space. To substantiate our claims, we use \muxlsr{} speech encoder in combination with a pre-trained \labse{} text sentence encoder for cross-lingual speech-to-text translation retrieval, and \muxlsr{} alone for cross-lingual speech-to-speech translation retrieval. We highlight these applications by performing several cross-lingual text and speech translation retrieval tasks across several datasets.
\end{abstract}


\section{Introduction}
Recently, self-supervised pre-training of large transformer encoders on massive amounts of unlabeled audio data followed by task-specific fine-tuning has emerged as the de-facto approach for achieving state-of-the-art performance on several tasks in spoken language processing. However, popular self-supervised representation learning (SSL) approaches such as Wav2vec-2.0 \cite{Baevski2020wav2vec} and others \cite{chung2020generative, npc, pase,schneider2019, khurana2020convolutional,conneau2020unsupervised,hsu2021hubert,babu2021xlsr,chen2021wavlm,w2v_bert,bapna2022mslam} learn speech embedding at acoustic frame-level, i.e., for short speech segments of duration 10 to 20 milliseconds.
\begin{figure}
    \centering
    \caption{An illustration of the cross-lingual multimodal embedding space.}
    \label{fig:vsm}
    \includegraphics[width=0.98\linewidth]{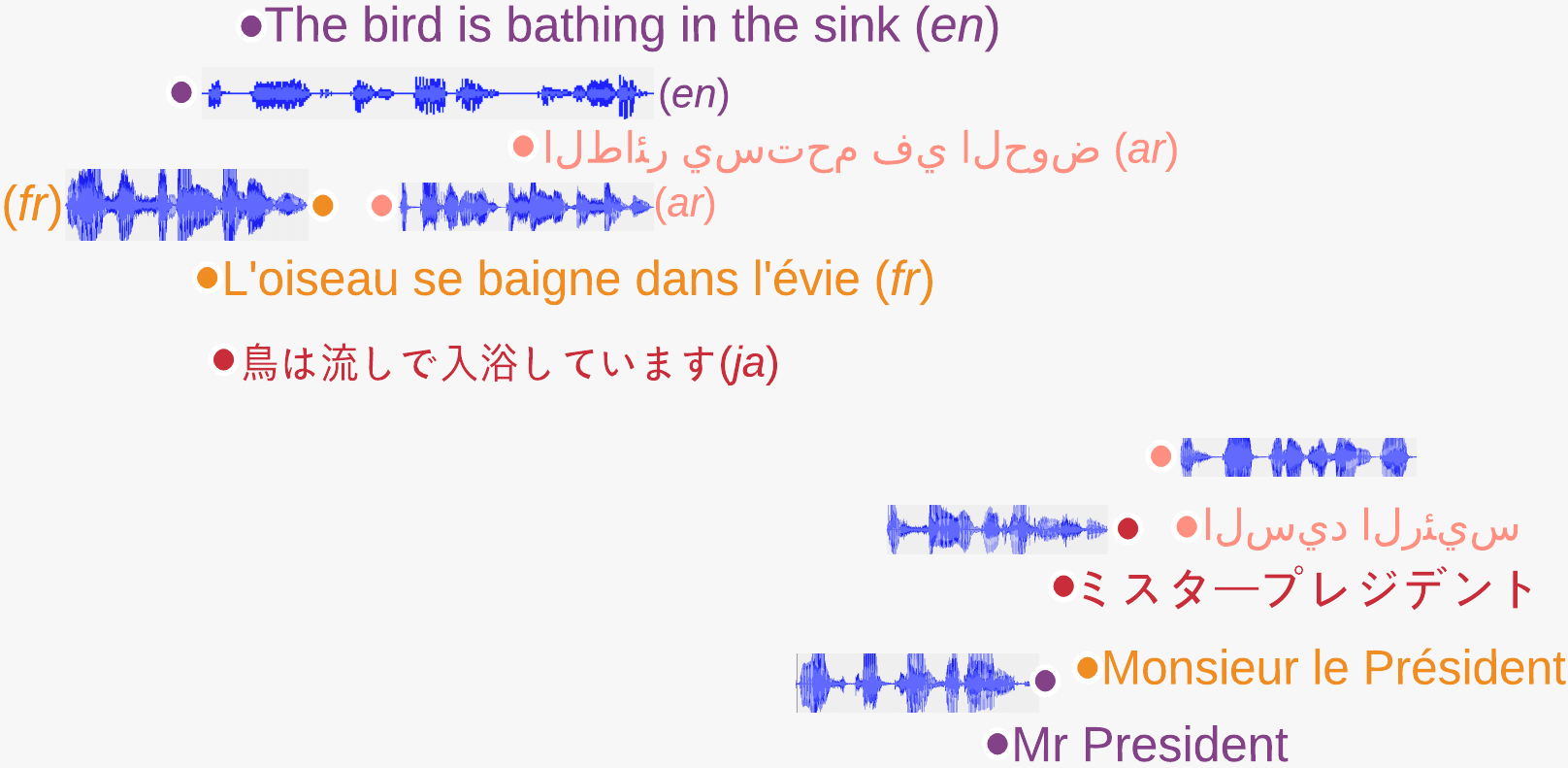}
\end{figure}

Unlike previous works mentioned above, this work focuses on learning semantically-aligned multimodal utterance-level cross-lingual speech representations (\muxlsr{}). The \muxlsr{}'s embedding vector space is multimodal since it is shared between the speech and the text modalities. It is cross-lingual since various languages share it. Furthermore, it's semantically aligned since, in the \muxlsr{}'s vector space, a spoken utterance is clustered together with its speech and text translations. We show a two-dimensional illustration of the desired embedding vector space in Figure~\ref{fig:vsm}. As an example, consider the English phrase \textit{A bird is bathing in the sink}. Now, in \muxlsr{}'s embedding space, the written form of the above phrase should be clustered together with its written and spoken forms in various languages (Japanese, French, and Arabic in the figure). And, in some other regions of the embedding space, the phrase \textit{Mr President} is clustered with its written and spoken form in several languages. Unfortunately, the acoustic frame-level unimodal contextual representation learning frameworks like Wav2vec-2.0 \cite{Baevski2020wav2vec} or the multilingual \XLSR{} \cite{conneau2020unsupervised, babu2021xlsr} do not learn an embedding space with the same properties. We believe that encoding semantics is one of the many missing pieces in the self-supervised speech representation learning puzzle. 

On the other hand, several transformer encoders for text have been proposed in recent years that go beyond token-level contextual representations and learn cross-lingual semantically-aligned sentence embedding vector spaces across several languages \cite{laser_old, Artetxe_2019,feng2020languageagnostic}. These models have found use in bi-text data mining. The task is to retrieve the text translation in a target language for a given sentence query in a source language by matching the query sentence embedding with those of sentences in the target language search database \cite{fair_mining, wiki_matrix, cc_matrix}. Given that text encoders can successfully learn semantically aligned cross-lingual sentence embedding spaces, we ask whether it is possible to make these text embedding spaces multimodal by learning to map speech utterances in the semantically-aligned cross-lingual text embedding space.

\begin{figure*}
    \centering
    \caption{A pedagogical description of how learning with transcribed speech data using \labse{} as the teacher could lead to the emergence of cross-lingual speech and text associations. In this illustration, we use English speech $x^{(\text{EN})}$ and its transcription $y^{(\text{EN})}$ for training. \muxlsr{}'s parameters are tuned to close the distance between the speech embedding given by \muxlsr{} in orange and \labse{}'s embedding (Anchor) of the corresponding text transcript in green. Since \labse{}'s text embedding space is semantically-aligned across various languages, by pulling the speech embedding towards the anchor embedding, we automatically learn cross-lingual speech-text alignments without ever seeing cross-lingual associations during training. In practice, we train \muxlsr{} with multilingual transcribed speech, not just English.}
    \label{fig:pedag}
    \includegraphics[width=0.7\linewidth]{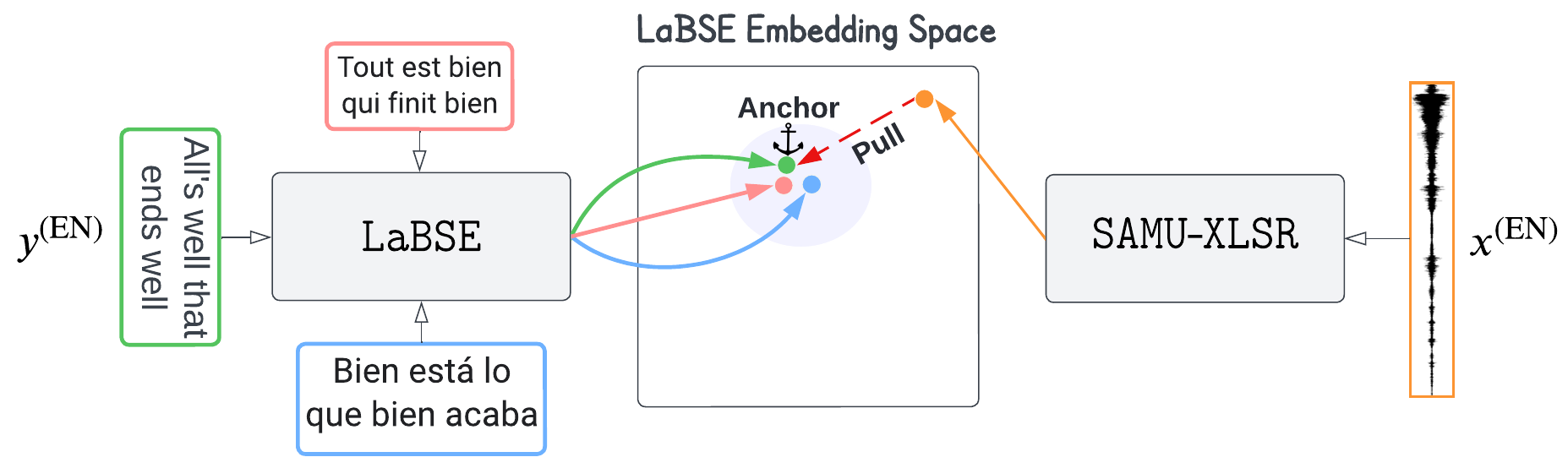}
\end{figure*}

To that end, we propose a multimodal learning framework for fine-tuning the pre-trained multilingual \XLSR{} speech encoder via knowledge distillation from the pre-trained language-agnostic BERT sentence encoder \labse{} \cite{feng2020languageagnostic}. Also, we append a pooling mechanism and a non-linear projection layer after the last layer of the pre-trained \XLSR{} encoder to transform the frame-level contextual representations into a single utterance level embedding vector. Then, we train the speech encoder using transcribed speech; given a speech utterance, the parameters of the speech encoder are tuned to accurately predict the text embedding provided by the \labse{} encoder of its corresponding transcript. Because \labse{}'s embedding vector space is semantically-aligned across various languages, the text transcript would be clustered together with its text translations. Hence, we get cross-lingual speech-to-text associations for free by simply using transcribed speech to train the speech encoder via the proposed knowledge distillation framework. For a pedagogical description, see Figure~\ref{fig:pedag}.

One of the use cases of the \muxlsr{} embedding space described above is for data mining. Recent years have seen remarkable progress in Automatic Speech Recognition across several domains and languages. The next frontier in spoken language processing is automatic speech to text and speech to speech machine translation. Developing speech-based MT systems would require massive amounts of parallel translated speech data in several languages, which could be highly costly to collect. But, the multimodal cross-lingual embedding space illustrated in Fig.~\ref{fig:vsm} could address this issue. We could build a cross-lingual speech to text and speech to speech retrieval pipeline, which could entirely or, in some cases, partially automate the process of collecting either text or speech translations corresponding to a spoken utterance. We advise the reader to look at papers in Natural Language Processing that use multilingual sentence encoders to perform cross-lingual text mining, such as \cite{schwenk2017learning, shwenk2019, schwenk2020ccmatrix, feng2020languageagnostic}.

Cross-lingual speech-to-text mining to create parallel speech-text translation datasets is just one possible application of \muxlsr{}. But, what motivates us to work on this problem is the potential application in zero-shot speech-to-text translation. The success of zero-shot translation depends on learning a semantically-aligned language invariant embedding vector space or an interlingua for different spoken languages, where speech utterances and their speech translations are clustered together. We show that this is an emergent property in \muxlsr{}'s embedding vector space as a result of training \muxlsr{} using the proposed multimodal learning framework (Section~\ref{sec:st}). Some of the text machine translation papers that inspire us in the field of zero-shot translation are \cite{gu-etal-2019-improved, arivazhagan2019missing}.

Through this work, we make the following \textbf{contributions}:
\begin{itemize}
    \item We propose a simple yet effective multimodal learning framework for semantically-aligned multimodal (joint speech-text) utterance-level speech representation (\muxlsr{}) shared across multiple languages (Section~\ref{sec:method}).
    \item First, we demonstrate the effectiveness of our models on several zero-shot cross-lingual speech-to-text and speech-to-speech translation retrieval tasks (Section~\ref{sec:results_discuss}).
    \item Second, we show that \muxlsr{} could be used for sequence-to-sequence modeling tasks such as phoneme recognition and Automatic Speech Recognition (ASR) (Section~\ref{sec:per}).
    \item Finally, we conduct analysis to understand better the various design decisions that went into constructing \muxlsr{} (Section~\ref{sec:anal}).
\end{itemize}
A work that is similar to ours is presented in \cite{duquenne2021multimodal}. Unlike the previous work, we evaluate our model on multiple datasets across many languages with a special emphasis on low-resource languages. 

Furthermore, unlike the multimodal speech encoder presented in \cite{duquenne2021multimodal}, we show that \muxlsr{} performs at par or better than \XLSR{} on the downstream ASR task across different languages. We recommend the reader to read \cite{duquenne2021multimodal} along with this paper to get a holistic understanding of this field. 
\section{Methodology}
\label{sec:method}
\subsection{Problem Formulation}
\begin{figure*}
    \centering
    \caption{An illustration of the multimodal training framework}
    \label{fig:1}
    \includegraphics[width=0.9\textwidth]{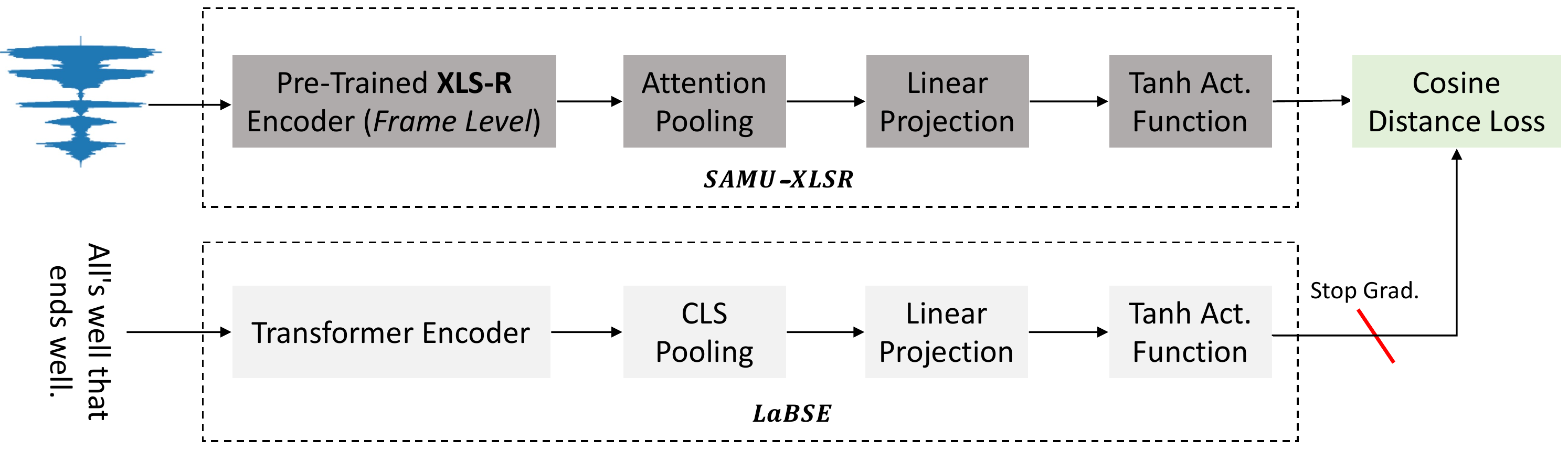}
\end{figure*}

We train \muxlsr{} using a multilingual set $\mathcal{D}$ of paired examples $(x^{(l)}, y^{(l)})$, where $x^{(l)}$ is the speech waveform, and $y^{(l)}$ is its text transcript in language $l$. Given a training example, $(x^{(l)}, y^{(l)})$, we transform the sequence of discrete tokens $y^{(l)}$ to a dense embedding vector $\mathbf{z}_T \in \mathbb{R}^{d}$ using a text encoder $g_{\phi}$, and the series of speech samples $x^{(l)}$ into a dense embedding vector $\mathbf{z}_S \in \mathbb{R}^{d}$ using a speech encoder $f_{\theta}$. Then, we update the parameters of the speech encoder $f_{\theta}$ so that the distance between the speech embedding $\mathbf{z}_S$ and the text embedding $\mathbf{z}_T$ is minimized. The training loss for a single example is given by the following equation:
\begin{equation}\label{eq:1}
\mathcal{J}(\theta, \phi) = \text{distance}(\mathbf{z}_S, \mathbf{z}_T)
\end{equation}
We use the pre-trained Language-agnostic BERT Sentence Encoder (\labse{}) as the text encoder $g_\phi$ and \muxlsr{} as the speech encoder $f_\theta$. The parameters $\theta$ of the speech encoder are updated during training, while the parameters $\phi$ of the text encoder remain fixed. An illustration of the multimodal learning framework is shown in Figure~\ref{fig:1}.

\subsection{\muxlsr{} Speech Encoder, $f_\theta$}
\label{sec:samuxlsr}
\muxlsr{} consists of a pre-trained frame-level \XLSR{} speech encoder \cite{babu2021xlsr} followed by a mechanism for pooling the frame-level contextual representations into a single embedding vector.

The \XLSR{} speech encoder consists of a deep convolutional neural network that maps 1D time series representing the sample values of the speech waveform into a 2D sequence of feature vectors $\mathcal{H} \in \mathbb{R}^{T \times 512}$. Each feature vector $h_t \in \mathcal{H}$ represents 20ms of the speech signal. The time resolution of $h_t$ is similar to that of an acoustic frame. Therefore, we refer to $\mathcal{H}$ as frame-level representations. Next, the feature sequence $\mathcal{H}$ is transformed into contextual representations $\mathcal{C} \in \mathbb{R}^{T\times 1024}$ by a deep transformer encoder \cite{vaswani2017attention}. The transformer encoder consists of 24 Multi-Headed Self-Attention (MHSA) transformer blocks. The attention vector size is 1024, and there are 16 attention heads in each block. We use the publicly available pre-trained \XLSR{} checkpoint\footnote{\url{https://huggingface.co/facebook/wav2vec2-xls-r-300m}} which was trained on 400k hours of unlabeled speech data in 128 languages.

Next, we use Self-Attention pooling \cite{safari2020self} strategy to get a single utterance-level embedding vector $\mathbf{e} \in \mathbb{R}^{1024}$. In this pooling strategy, we take a weighted combination $\sum\limits_{t=1}^{T} v_t c_t$ of contextual vectors $c_t \in \mathcal{C}$, where $\mathbf{v} = (v_1, \ldots, v_T)$ is the attention vector, given by the following equation:
\begin{equation}
    \mathbf{v} = \text{softmax} (\mathcal{C}\mathbf{w})
\end{equation}
where, $\mathbf{w}\in \mathbb{R}^{1024}$, which gives $\mathbf{v} \in \mathbb{R}^{T}$, such that $\sum\limits_{t} v_t = 1$. The weight vector $\mathbf{w}$ is learned during training.

Finally, we take a non-linear projection of the embedding vector $\mathbf{e}$ to get the speech embedding $\mathbf{z}_S$. Overall, the \muxlsr{} speech encoder consists of approximately 300 million trainable parameters (weights and biases).
\subsection{LaBSE Text Encoder, $g_\phi$}
The key ingredient in our proposed multimodal learning framework is the \labse{} text encoder $g_\phi$, which allows us to learn a joint speech-text embedding space that is semantically aligned and shared across different languages. \labse{} is a language-agnostic text encoder for text with an architecture similar to the BERT transformer encoder \cite{devlin2019bert}. However, unlike BERT, \labse{} is a sentence embedding model, which is trained using both masked \cite{devlin2019bert} and translation language modeling \cite{lample2019cross} objective functions. \labse{} consists of a token level transformer encoder with 12 MHSA layers, followed by a pooling mechanism to construct a dense sentence-level embedding vector. 

The \labse{}'s transformer encoder takes as input text that is tokenized into "wordpieces" \cite{wordpieces_shuster, wordpieces} and outputs a sequence of contextual token embedding $\mathcal{W} \in \mathbb{R}^{L\times 768}$. A non-linear projection of the CLS token embedding is used as the sentence embedding $\mathbf{z}_T \in \mathbb{R}^{768}$, which is used as the training target for \muxlsr{} training. We use the pre-trained \labse{} model checkpoint\footnote{\url{https://huggingface.co/sentence-transformers/LaBSE}} hosted on the Huggingface \cite{hugging} models\footnote{\url{https://huggingface.co/models}} platform. We refer to the use of CLS token embedding for sentence representation as CLS pooling to conform with the terminology used in the Huggingface hosted \labse{} encoder.

\labse{} embeds sentences from 109 languages into a shared semantically-aligned embedding vector space. Unlike \labse{}, other multilingual text encoders such as XLM-R \cite{xlmr} do not learn an aligned sentence embedding space. Therefore, to achieve our goal of embedding speech in a semantically aligned vector space, we use \labse{} as the teacher for training \muxlsr{}.

\subsection{\muxlsr{} Training Details}\label{sec:train}
\subsubsection{Training Data, $\mathcal{D}$}\label{subsec:d} We train \muxlsr{} on transcribed speech in 25 languages derived from the publicly available CommonVoice-v7 (CoVo) dataset. The 25 languages are namely, English (EN), French (FR), German (DE), Spanish (ES), Catalan (CA), Italian (IT), Welsh (CY), Russian (RU), Chinese (China) (ZH\_CN), Chinese (Taiwan) (ZH\_TW), Chinese (Hong Kong) (ZH\_HK), Portuguese (PT), Polish (PL), Persian (FA), Estonian (ET), Mongolian (MN), Dutch (NL), Turkish (TR), Arabic (AR), Swedish (SV\_SE), Latvian (LV), Slovenian (SL), Tamil (TA), Japanese (JA) and Indonesian (ID). Table~\ref{tab:covo_data} shows the per-language transcribed data available in CoVo. The total training data size is 6.8K hours. 

\begin{figure}
    \caption{Re-balancing the training set with different values of the smoothing parameter $\alpha$}
    \label{fig:alphas}
    \centering
    \includegraphics[width=0.99\linewidth]{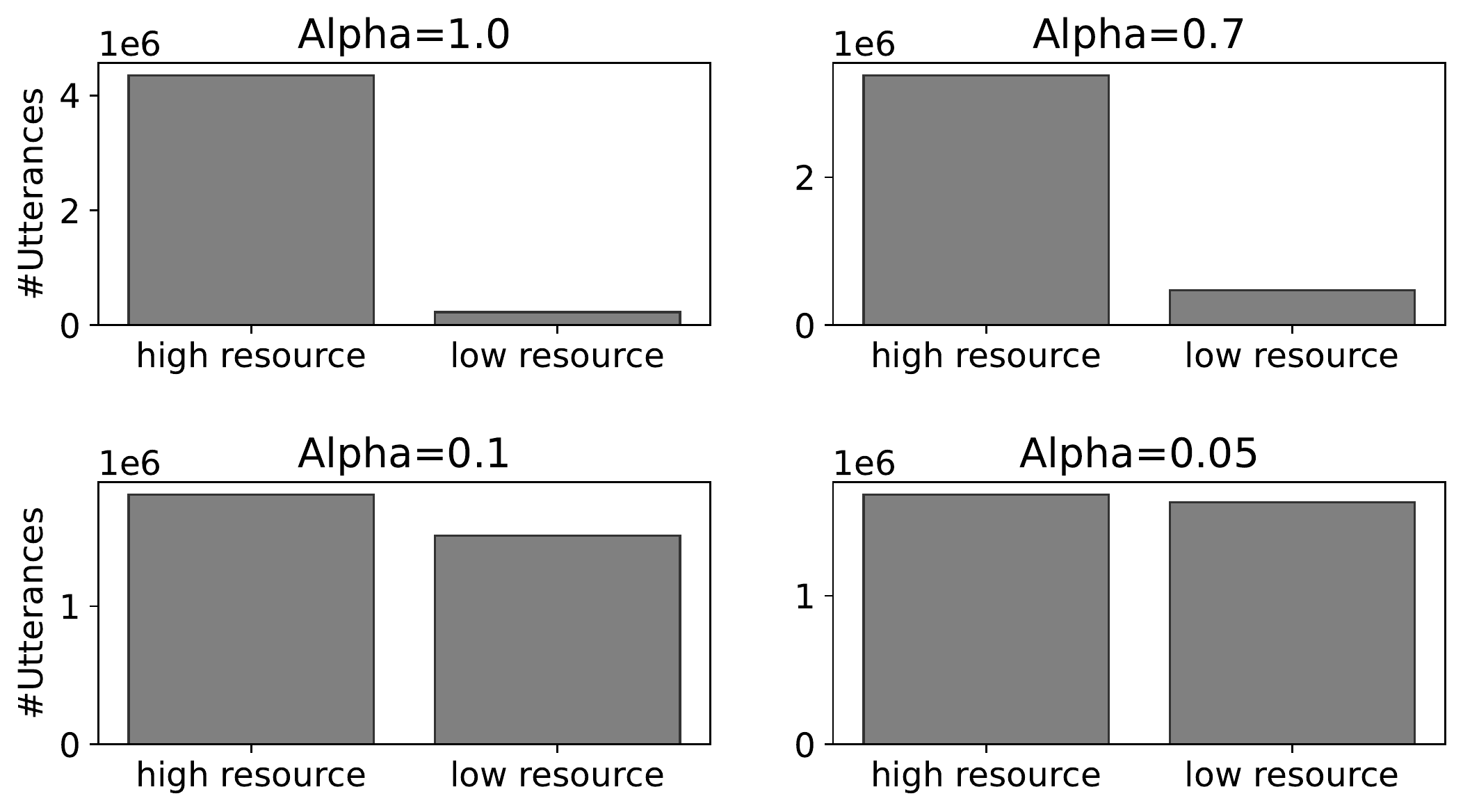}
\end{figure}

Clearly, the data is highly imbalanced. The top 5 high-resource languages make up 72\% of the training data, while the bottom 14 low-resource languages make up just 10\% of the training data. The above mentioned problem could lead to \muxlsr{} severely under-fitting on low-resource languages, because \muxlsr{}, during its training lifetime, might encounter transcribed speech data from low-resource languages in its train mini-batch only a few times. Following \cite{rebalance_data, rebalance_data_v2} we re-balance the training set $\mathcal{D}$ by up/down-sampling data from each language $l$ with a ratio $\lambda_l$:
\begin{equation}
\label{eq:2}
    \lambda_l = \frac{1}{p_l} \frac{p_l^{\alpha}}{\sum\limits_l p_l^{\alpha}}\ \text{with}\ p_l = \frac{n_l}{\sum\limits_{l=1}^{L} n_l}
\end{equation}
where, $\alpha$ is the smoothing parameter, $n_l$ is the number of utterances for language $l$ in the training set. Figure~\ref{fig:alphas}, shows how varying $\alpha$ between 1.0 and 0.05 re-balances the training set. As we make $\alpha$ smaller, observe that the share of low-resource languages in the training set becomes approximately same as that of high-resource languages. It is important to note that when we up-sample data from low-resource languages, we simply repeat the utterances from those languages, and, down-sampling data from high-resource languages involve picking random utterances according to the ratio $\lambda_l$. Hence, training with a re-balanced training set that is created using a small value of $\alpha$ could result in a drop in performance on high-resource languages as compared to the model that is trained with the original unbalanced training set. We study the effect that the smoothing parameter $\alpha$ has on the model's downstream task performance in Section~\ref{sec:alpha}.

\subsubsection{Optimization Settings} We train \muxlsr{} for 400K training iterations, on 32 V100-32gb GPUs, with a per-GPU mini-batch size of approximately 2 hours of transcribed speech. Following \cite{conneau2020unsupervised}, we use the Adam optimizer for updating the model parameters with a three phase learning rate scheduler; Warm-up the learning rate to a maximum value of 1e-4 for the first 10\% of the training iterations, then the learning rate remains constant for the next 40\% of the training iterations, and finally decays linearly for the rest of the iterations. For the first 10K training iterations, only the projection layer of \muxlsr{} encoder is trained while the pre-trained frame-level \XLSR{} speech encoder remains fixed. We do not update the weights of the \XLSR{}'s convolutional feature extractor throughout the training process. Also, we use a modified version of SpecAugment \cite{park2019specaugment} on the feature sequence $\mathcal{H}$ (Section~\ref{sec:samuxlsr}) to mask the input to the \XLSR{}'s transformer encoder, which leads to better performance on downstream tasks. The above mentioned training settings are the standard for fine-tuning the pre-trained \XLSR{} or wav2vec-2.0 speech encoders on downstream ASR tasks \cite{Baevski2020wav2vec, conneau2020unsupervised}.
\begin{table}[]
    \caption{Amount of per language transcribed speech data in the CommonVoice-v7 dataset}
    \label{tab:covo_data}
    \centering
    \renewcommand{\arraystretch}{1.3}
    \resizebox{.8\linewidth}{!}{%
    \begin{tabular}{llllll}
\toprule
     \textbf{Lang} & \textbf{EN} &  \textbf{DE} &  \textbf{CA} &  \textbf{FR} &  \textbf{ES} \\
\textbf{Dur [Hrs]} & 2K & 960 & 790 & 740 & 380 \\\midrule
     \textbf{Lang} &  \textbf{FA} &  \textbf{IT} &  \textbf{CY} &  \textbf{TA} &  \textbf{RU} \\
\textbf{Dur [Hrs]} & 290 & 290 & 220 & 200 & 150 \\\midrule
     \textbf{Lang} &  \textbf{PL} & \textbf{ZH\_HK} & \textbf{NL} & \textbf{PT} & \textbf{AR} \\
\textbf{Dur [Hrs]} & 130 &    96 & 93 & 85 & 84 \\\midrule
     \textbf{Lang} & \textbf{ZH\_CN} & \textbf{ZH\_TW} & \textbf{SV\_SE} & \textbf{ET} & \textbf{TR} \\
\textbf{Dur [Hrs]} &    63 &    59 &    34 & 32 & 32 \\\midrule
     \textbf{Lang} & \textbf{JA} & \textbf{ID} & \textbf{MN} & \textbf{SL} & \textbf{LV} \\
\textbf{Dur [Hrs]} & 27 & 25 & 12 &  9 &  7 \\
\end{tabular}
}
\end{table}

We use the cosine distance between the speech and the text embedding as the training loss (Equation~\ref{eq:1}). We do not update the weights of the \labse{} text encoder throughout training. The reason for this design choice is straightforward. \labse{}'s sentence embedding space is already semantically aligned across 109 languages. By fine-tuning \labse{} along with \muxlsr{} on transcribed speech data $\mathcal{D}$, we run the risk of destroying this alignment. In fact, \labse{} will have no incentive to maintain an aligned embedding space. Instead, our learning framework simply attempts to embed speech utterances in the \labse{}'s sentence embedding space to make it multimodal. By simply forcing the speech embeddings outputted by \muxlsr{} to be closer to \labse{} text embedding, we get the cross-lingual semantic alignments between speech utterances in different languages and text in 109 languages without ever encountering cross-lingual associations during the model's training. Having said that, it might be possible to train the \labse{} text encoder along with \muxlsr{} and still maintain the \labse{}'s semantically aligned embedding space. But, it is out-of-scope of this paper.

\subsection{\muxlsr{} Model Card}
Table~\ref{tab:card} summarizes the best configuration of different hyperparameters for training \muxlsr{} encoder. Next, we explain what some parameters in the table mean. CoVo\_25 refers to the multilingual transcribed speech data used for training the model. We use data in 25 languages from the CoVo dataset. CNN Feature Extractor refers to the pre-trained \XLSR{}'s convolutional encoder that maps the 1D speech waveform to a 2D feature representation that is used as input to the transformer encoder. We keep its weights fixed to the pre-trained value. Freeze Fine-tune updates refer to the number of training iterations up to which we only train the projection layer of \muxlsr{}. See Equation~\ref{eq:2} and the text above it for details on the smoothing factor $\alpha$. The learning rate scheduler (LR scheduler) has a value of 10-40-50 refers to the learning rate scheduler mentioned in Section~\ref{sec:train}. Training teacher is \labse{} which refers to the fact that the training targets for \muxlsr{} are the embedding vectors corresponding to the text transcripts provided by \labse{}. The model supports 25 spoken languages and 109 written languages since \muxlsr{} is trained on the transcribed speech from 25 languages and \labse{} can encode text in 109 languages in its semantically aligned cross-lingual vector space. 

\begin{table}[]
    \caption{\muxlsr{} model card}
    \label{tab:card}
    \centering
    \renewcommand{\arraystretch}{1.3}
    \resizebox{.99\linewidth}{!}{%
    \begin{tabular}{ll}
\toprule
 \textbf{Parameters} & \textbf{Value}\\\midrule
 Training Data & CoVo\_25 \\
 Smoothing factor ($\alpha$) for data re-balancing & 0.05 \\
 Training updates & 200K \\
 Freeze Fine-tune updates & 10k \\
 CNN Feature Extractor & Frozen \\
 Optimizer & Adam\\
 max learning rate (LR) & 1e-4\\
 LR scheduler & 10-40-50\\
 batch size / GPU & 2Hrs \\
 Data Augmentation & SpecAugment on $\mathcal{H}$ \\
 Training Objf. & Cosine Distance\\
 Training Teacher & \labse{} \\
 Pooling Fn. & Self-Attention \\
 Model init. & XLSR Pre-Trained checkpoint \\
 Num. GPUs & 32 \\
 Supported Spoken Langs & 22 \\
 Supported text Langs & 109 \\
\end{tabular}}
\end{table}

\section{Downstream Evaluation Tasks \& Metrics}\label{sec:tasks}
\subsection{Overview}
\textbf{Retrieval}: We evaluate our multimodal framework (Fig.~\ref{fig:1}) that consists of \muxlsr{}, a speech embedding model, and \labse{}, a text embedding model, on several downstream translation retrieval tasks. Retrieval is a common way to evaluate multilingual semantically aligned sentence embedding vector spaces in Natural language processing \cite{schwenk2017learning,feng2020languageagnostic}.

As mentioned before, our work aims to learn a semantically aligned cross-lingual multimodal (joint speech-text) embedding space. Hence, if successful at achieving our desired goal, the \muxlsr{}-\labse{} combination should give good performance on cross-lingual speech-to-text translation retrieval tasks. Also, \muxlsr{} alone should be able to perform well on cross-lingual speech-to-speech translation retrieval tasks.

\textbf{Sequence Generation}: Furthermore, we perform sequence-to-sequence modeling tasks, namely the Connectionist Temporal Classification (CTC) \cite{Graves12} based Phoneme Recognition (generating the underlying phoneme sequence corresponding to an input speech sequence) and Automatic Speech Recognition (ASR) (generating the underlying word sequence corresponding to an input speech sequence) using \muxlsr{}.

\subsection{Translation Retrieval Tasks}
\label{sec:retrieval_tasks}
Here, we summarize the retrieval process, evaluation metrics and the speech-to-text and speech-to-speech translation retrieval tasks we use to evaluate the  \muxlsr{}'s multimodal semantic embedding space.

\textbf{Retrieval process and Evaluation Metrics}: We construct two databases (DB), query and search, to perform translation retrieval. The query DB consists of speech utterances in a language X, and in the case of text translation retrieval tasks, the search DB consists of text sentences in a language Y. The task is to retrieve the correct text translation from the search DB corresponding to each speech query in the query DB. To that end, we transform the speech utterances in the query DB through \muxlsr{} to query speech embedding matrix $Q\in \mathbb{R}^{N\times 768}$, where $N$ is the number of speech queries in the query DB. Also, we transform the sentences in the search DB through the \labse{} encoder to search text embedding matrix $S\in \mathbb{R}^{M\times 768}$, where $M$ is the number of sentences in the search DB. Given that the vectors are normalized, we could retrieve the text translations for the speech queries as follows:
\begin{align*}
    A &= QS^{T}\\
    \mathbf{r} &= \text{argmax}_{j} A_{:,j}
\end{align*}
where, $A\in \mathbb{R}^{N\times M}$ is the cosine similarity matrix, whose $(i, j)^{th}$ element $A_{i,j}$ is the cosine similarity between the speech query embedding $q_i \in Q$ and the sentence embedding $s_j \in S$, and $\mathbf{r} \in \mathbb{R}^{N}$ is the index vector, such that its each component $r_{i} \in \mathbf{r}$ is the index of the closest match in the text translation search DB. Also, given the index vector $\mathbf{u}$, where each component $u_j \in \mathbf{u}$ is the index of the ground-truth text translation in the search DB, we compute the model's retrieval accuracy as follows:
\begin{equation}\label{eq:acc}
    \text{ACC} = 100*\frac{\sum\limits_{i=1}^{N} 1\{r_i = u_i\}}{N}
\end{equation}
where, the function $1\{r_i = u_i\}$ returns one when $r_i = u_i$, the predicted translation index matches the ground-truth translation index, otherwise it outputs zero. Hence, the numerator is the number of queries for which the model retrieved the correct translations from the search DB and the denominator is the total number of queries in the query DB. 

We refer to the retrieval accuracy in Equation~\ref{eq:acc} as Recall@1 or R@1, which contrasts with another similar metric, R@5, where the indicator function returns one if any of the top five retrieved search DB indices matches with the correct index. We report R@5 for speech retrieval evaluation tasks. The recall is commonly used to evaluate audio-visual multimodal representation learning models \cite{harwath2016unsupervised, Harwath2020Learning, avlnet}. 

In addition to R@1, for text translation retrieval tasks, we also report the Word Error Rate (WER) \cite{enwiki:939575741} between the retrieved and the ground-truth text translation. The reason is that it is hard to interpret retrieval accuracies. For example, WER for model A with a retrieval accuracy of 70\% might not be much worse than the WER for model B with a retrieval accuracy of 80\% because model A might be worse than model B in retrieving the exact translations. However, it might still recover translations with a significant string overlap with the actual translation. The retrieval accuracy will fail to capture this.

\textbf{X$\rightarrow$EN Text Translation Retrieval}: We use the CoVoST-2 \cite{wang2020covost} X-EN speech-translation dataset for this evaluation task. The speech query DB is in a language X$\in$\{RU, IT, FR, ES, TR, DE, ET, CY, NL, ID, CA, FA, AR, ZH, SV, MN, SL, JA, TA, LV\} and the search DB consists of English sentences. To construct the speech query DB for each language X, we use the combined testing and development sets (henceforth, eval set) from CoVoST-2. To construct the search DB, we combine the English text translation from all the 22 X$\rightarrow$EN eval sets in CoVoST-2, which we refer to as $S_a$. In addition, we create a search DB $S_b$, that contains approximately 1.4M English sentences from the CoVo English transcribed speech data. We use the combined search DB $S = S_a\cup S_b$ for all the 22 X$\rightarrow$EN text translation retrieval tasks. We add $S_b$ to $S_a$ to make the retrieval task harder than if we just search over $S_a$.

\textbf{EN$\rightarrow$Y Text Translation Retrieval}: We use the the publicly available CoVoST-2 corpora \cite{wang2020covost} for this evaluation task, which consists of English speech queries paired with their text translations. The speech query DB is in English and search DB is in a language Y$\in$\{DE, CA, ZH, FA, ET, MN, TR, AR, SV, LV, SL, TA, JA, ID, CY\}. For each EN$\rightarrow$Y retrieval task, the query DB consist of speech utterances in the combined development and testing sets. The search DB consists of the true text translations in language Y. corresponding to the speech queries. In addition, we add the Y language text translations available in the EN$\rightarrow$Y CoVoST-2 training set to make the retrieval task harder. Similarly, we create a search DB for each of the 15 languages Y for the EN$\rightarrow$Y text translation retrieval task.

For this evaluation scenario, we also perform text translation retrieval on the MUST-C \cite{di-gangi-etal-2019-must} EN$\rightarrow$Y corpora. In MUST-C, we have English speech queries paired with their true text translation in a language Y$\in$\{ES, PT, FR, DE, Romanian (RO), NL, IT, Czech (CS), Vietnamese (VI), FA, TR, AR, RU, ZH\}. We create an eval set, a union of MUST-C dev, tst-COMMON and tst-HE data splits. The speech query DB consists of speech utterances in the eval set. The search DB for a language Y consists of sentences from the EN$\rightarrow$Y MUST-C eval set combined with sentences from the EN$\rightarrow$Y training set.

\textbf{X$\rightarrow$Y Text Translation Retrieval}: We use the MTEDx \cite{salesky2021multilingual} speech-translation corpora, which consists of speech queries in language X paired with their ground-truth text translation. For this evaluation task, we have the translation pairs X\_Y$\in$\{IT\_ES, IT\_EN, ES\_FR, ES\_IT, FR\_PT, ES\_PT, FR\_EN, PT\_ES, ES\_EN, PT\_EN, RU\_EN\}. For a translation pair X\_Y, we have speech queries in language X and the text search DB in language Y. For a retrieval X$\rightarrow$Y, the query DB consists of speech utterances in the MTEDx X$\rightarrow$Y eval set (dev+test), and the text search DB in language consists of the ground-truth text translations from the X$\rightarrow$Y eval set and the X$\rightarrow$Y training set. The reader might observe that the search DB is more significant than the query DB for all the text translation retrieval tasks and consists of the actual text translations and random sentences to make the retrieval task harder.
\begin{table*}[]
    \centering
    \caption{We perform \textbf{zero-shot} X$\rightarrow$EN text translation retrieval on \textbf{In-domain} CoVoST-2 dataset. The search database for all X$\rightarrow$EN retrieval tasks consists of 1.6 million English sentences. We give the number of speech utterances in the query database for each retrieval task below. The task is to retrieve the correct text translation for the speech queries in language X. We report the Retrieval accuracy (R@1) and the Word Error Rate between the ground-truth and retrieved text translations. We compare our retrieval pipeline \muxlsr{}-\labse{}, with \asr{}-\labse{} and the Topline retrieval model. The \muxlsr{}-\labse{} retrieval pipeline transforms speech queries to embedding vectors using our \muxlsr{} speech encoder. Then, we match the query embedding vectors with the \labse{} text embeddings of the sentences in the search DB to retrieve the translation. The \asr{}-\labse{} retrieval pipeline first uses an ASR for language X to transcribe speech queries and then uses \labse{} to perform text-to-text translation retrieval. The Topline model uses the ground-truth text transcripts for the speech queries and performs text-to-text translation retrieval tasks using \labse{}.}
    \label{tab:x_en_res}
    \renewcommand{\arraystretch}{1.3}
    \resizebox{.99\linewidth}{!}{%
   \begin{tabular}{lrrrrrrrrrrrrrrrrrrrr}
\toprule
 \textbf{X} &   \textbf{RU} &   \textbf{IT} &   \textbf{FR} &   \textbf{ES} &   \textbf{TR} &   \textbf{DE} &   \textbf{ET} &   \textbf{CY} &   \textbf{NL} &   \textbf{ID} &   \textbf{CA} &   \textbf{FA} &   \textbf{AR} &  \textbf{ZH} &  \textbf{SV} &   \textbf{MN} &   \textbf{SL} &   \textbf{JA} &   \textbf{TA} &  \textbf{Avg.} \\
\midrule
\textbf{Query DB} &12K&18K&30K&26K&3.3K&27K&3.1K&1.4K&3.4K&1.6K&25K&6.8K&3.5K&9.7K&2.9K&3.5K&870&1.3K&1.2K&-\\\midrule
\multicolumn{21}{l}{\muxlsr{}-\labse{} Speech(X)$\rightarrow$Text(EN) Retrieval} \\\midrule
\textbf{R@1[\%]} & 93.5 & 92.9 & 92.5 & 92.9 & 93.4 & 90.9 & 91.5 & 84.6 & 89.7 & 84.4 & 82.1 & 83.6 & 73.7 &   78.6 &   72.4 & 68.2 & 52.1 & 48.9 & 42.4 &  76.8 \\
\textbf{WER[\%]} &  2.6 &  3.0 &  3.5 &  3.6 &  3.7 &  4.7 &  4.8 &  5.1 &  4.9 &  9.5 & 11.0 & 10.2 & 13.8 &   15.2 &   19.0 & 26.0 & 32.4 & 44.7 & 57.7 &  17.2 \\\midrule
\multicolumn{21}{l}{\asr{}-\labse{} Speech(X)$\rightarrow$Text(EN) Retrieval} \\\midrule
\textbf{R@1[\%]} & 92.7 & 90.1 & 90.4 & 91.3 & 90.9 & 88.2 & 94.8 & 81.7 & 89.3 & 65.6 & 80.6 & 76.1 & 54.0 & 55.4&63.9&53.9 &   64.0 & 23.6 & 26.5 &71.7 \\
\textbf{WER[\%]} &  3.0 &  4.8 &  5.0 &  4.6 &  5.8 &  6.5 &  2.1 &  7.6 &  5.3 & 23.4 & 11.5 & 16.8 & 34.3 &   36.0&17.2&41.3 &   16.7 & 72.9 & 75.0 &20.9 \\\midrule
\multicolumn{21}{l}{Topline \labse{} Text(X)$\rightarrow$Text(EN) Retrieval}\\\midrule
\textbf{R@1[\%]} & 94.4 & 94.0 & 94.8 & 94.3 & 94.2 & 93.2 & 97.5 & 86.2 & 90.8 & 91.3 & 83.8 & 85.1 & 74.5 &   81.4 &   87.0 & 81.3 & 70.9 & 83.1 & 49.2  &  85.2 \\
\textbf{WER[\%]} &  2.0 &  2.5 &  1.9 &  2.6 &  2.9 &  2.8 &  0.4 &  4.1 &  4.2 &  2.5 &  9.9 &  8.7 & 13.5 &   12.8 &    4.7 & 14.4 & 10.2 &  9.4 & 51.7 & 8.7 \\
\end{tabular}}
\end{table*}

We consider MTEDx X$\rightarrow$Y translation retrieval evaluation tasks as out-of-domain because we train \muxlsr{} on transcribed read speech from the CoVo dataset. At the same time, MTEDx consists of oratory speech collected from TED talks.

\textbf{X$\rightarrow$EN Speech Translation Retrieval}: Finally, we evaluate our model on speech translation retrieval tasks. We get the parallel X$\rightarrow$EN speech-speech translation data from the publicly available VoxPopuli corpora \cite{wang-etal-2021-voxpopuli}. For this task, speech queries are in a language X$\in$\{ES, FR, PL, NL, DE, RO, Croatian (HR), CS\} and the search DB consists of English speech translations corresponding to the queries. Unlike the text translation retrieval tasks, the search DB is the same size as the query DB and consists of only actual speech translations corresponding to the queries.
\begin{table*}[]
    \centering
    \caption{We perform \textbf{zero-shot} EN$\rightarrow$Y text translation retrieval on \textbf{In-domain} CoVoST-2 dataset. The search database for each EN$\rightarrow$Y retrieval task consists of 320K sentences in language Y, and the query database consists of 31K English speech utterances. The task is to retrieve the correct text translation for the English speech queries. We report the Retrieval accuracy (R@1) and the Word Error Rate between the ground-truth and retrieved text translations. We compare our retrieval pipeline \muxlsr{}-\labse{}, with \asr{}-\labse{} and the Topline retrieval model. The \muxlsr{}-\labse{} retrieval pipeline transforms speech queries to embedding vectors using our \muxlsr{} speech encoder. Then, we match the query embedding vectors with the \labse{} text embeddings of the sentences in the search DB to retrieve the translation. The \asr{}-\labse{} retrieval pipeline first uses an English language ASR to transcribe speech queries and then uses \labse{} to perform text-to-text translation retrieval. The Topline model uses the ground-truth text transcripts for the speech queries and performs text-to-text translation retrieval tasks using \labse{}.}
    \label{tab:en_y_res}
    \renewcommand{\arraystretch}{1.3}
    \resizebox{.99\linewidth}{!}{%
    \begin{tabular}{lrrrrrrrrrrrrrrrr}
\toprule
\textbf{Y} &  \textbf{ZH} &   \textbf{SL} &   \textbf{TR} &   \textbf{LV} &   \textbf{CY} &   \textbf{ID} &   \textbf{DE} &   \textbf{CA} &   \textbf{AR} &  \textbf{SV} &   \textbf{ET} &   \textbf{TA} &   \textbf{FA} &   \textbf{JA} &   \textbf{MN} & \textbf{Avg.} \\\midrule
\multicolumn{17}{l}{\muxlsr{}-\labse{} Speech(EN)$\rightarrow$Text(Y) Retrieval}\\\midrule
 \textbf{R@1[\%]} &   87.2 & 90.5 & 89.4 & 89.9 & 90.8 & 91.0 & 91.5 & 91.4 & 88.3 &   91.7 & 90.4 & 90.5 & 89.0 & 88.1 & 86.2 &  89.7 \\
 \textbf{WER[\%]} &   11.2 &  6.3 &  7.4 &  7.2 &  6.2 &  5.9 &  5.8 &  5.5 &  8.5 &    5.5 &  6.6 &  7.3 &  8.4 & 11.9 & 10.9 &   7.6 \\\midrule
 \multicolumn{17}{l}{\asr{}-\labse{} Speech(EN)$\rightarrow$Text(Y) Retrieval}\\\midrule
 \textbf{R@1[\%]} &   87.9 & 90.6 & 89.8 & 90.2 & 90.7 & 91.2 & 91.4 & 91.6 & 89.0 &   91.7 & 90.5 & 91.2 & 89.6 & 88.4 & 87.3 &  90.1 \\
\textbf{WER[\%]} &   10.7 &  6.2 &  7.1 &  6.9 &  6.2 &  5.7 &  5.8 &  5.3 &  7.8 &    5.4 &  6.5 &  6.5 &  7.7 & 11.5 &  9.8 &   7.3 \\\midrule
 \multicolumn{17}{l}{Topline \labse{} Text(EN)$\rightarrow$Text(Y) Retrieval}\\\midrule
\textbf{R@1[\%]} &   95.8 & 97.1 & 96.2 & 96.6 & 96.7 & 96.8 & 97.1 & 96.9 & 95.7 &   97.3 & 96.7 & 97.0 & 95.4 & 95.5 & 94.5 &  96.4 \\
\textbf{WER[\%]} &    2.7 &  1.3 &  1.9 &  1.8 &  1.7 &  1.5 &  1.5 &  1.3 &  2.3 &    1.3 &  1.6 &  1.8 &  2.8 &  4.2 &  3.5 &   2.1\\
\end{tabular}}
\end{table*}
\subsection{Sequence-to-Sequence Modeling Tasks}\label{sec:seq_to_seq}
\textbf{Phoneme Recognition}: Phoneme recognition refers to the task of automatically decoding the underlying phoneme sequence $y$ corresponding to a speech sequence ($\mathbf{x}$). We Fine-tune the Pre-trained \muxlsr{} using paired $(\mathbf{x}, y)$ examples drawn from the CoVo dataset. Following \cite{rivire2020unsupervised, conneau2020unsupervised}, we build a phoneme recognizer for nine different languages, namely ES, FR, IT, Kabyle (KY), NL, RU, SV, TR, and Tatar (TT). We use one hour of transcribed data for training, 20mins for validation (model selection), and one hour for testing. The data splits are the same ones proposed in \cite{rivire2020unsupervised} and used in \cite{conneau2020unsupervised} for evaluating \XLSR{} on the phoneme recognition task. Our Fine-tuning setup matches the \XLSR{} Fine-tuning setup used in \cite{conneau2020unsupervised}.

\textbf{Automatic Speech Recognition}: ASR refers to the task of automatically decoding the underlying word sequence corresponding to a speech utterance. The Fine-tuning setup is the same as that for Phoneme Recognition. However, instead of phoneme sequence as the target for training, we have character sequences. To generate the word sequence from decoded character sequence, we use CTC beam search with a character-level N-gram language model.

We use the Espnet speech recognition toolkit \cite{watanabe2018espnet, arora2022espnet} for Fine-tuning the Pre-trained \muxlsr{} and \XLSR{ } models for sequence-to-sequence modeling tasks.

We believe that evaluating \muxlsr{} on sequence generation tasks mentioned above is interesting because it would be good to know whether \muxlsr{}, a speech encoder that we train using an utterance-level objective function (See Fig.~\ref{}), could also be used for tasks other than the utterance-level text and speech translation retrieval.

Another thing to note is that for sequence generation tasks, we dissect \muxlsr{} before the attention pooling layer (See Fig.~\ref{fig:1} to look at \muxlsr{}'s architecture) and use the computational modules before the pooling layer because for sequence generation tasks, we want a representation at the acoustic frame-level instead of the utterance level embedding outputted by \muxlsr{}. 

\section{Downstream Tasks: Zero-shot Translation Retrieval}
\label{sec:results_discuss}
\subsection{Additional Retrieval Models for comparison with \muxlsr{}}
\textbf{\asr{}-\labse{} retrieval pipeline}: We also perform translation retrieval tasks using an \asr{}-\labse{} combination, where we convert the speech queries into text transcripts in the same language as the queries using an ASR model. Then, we perform ASR transcript to text translation retrieval using \labse{}. We build 25 language-specific ASR models to cover all the spoken languages in our text translation retrieval tasks. To construct the ASR models, we fine-tune the pre-trained \XLSR{} checkpoint on the downstream ASR task using the transcribed speech data in the target language available from the CoVo dataset (See Table~\ref{tab:covo_data} for the amount of per language transcribed speech data). We use the standard Connectionist temporal Classification \cite{Graves2006} based optimization setup for fine-tuning the \XLSR{} model for the ASR task detailed in \cite{conneau2020unsupervised}. We use a beam size of 20 and a tri-gram character-level language model for decoding speech queries to text. We use the ESPnet speech recognition toolkit \cite{watanabe2018espnet, watanabe20212020} for constructing the ASR models and decoding.

\textbf{Topline}: As a topline, we use the ground-truth transcriptions corresponding to speech queries and perform ground-truth transcription to text translation retrieval using \labse{}. Our \muxlsr{}-\labse{} retrieval framework cannot perform better than the topline. Because the best we can do with our proposed multimodal learning framework is to match the \labse{} embedding vectors perfectly.


\subsection{Results on X$\rightarrow$EN text translation retrieval tasks} Table~\ref{tab:x_en_res} shows the results on X$\rightarrow$EN translation retrieval tasks using \muxlsr{}-\labse{}, \asr{}-\labse{} and Topline \labse{} retrieval pipelines. We report the retrieval accuracy (R@1) and WERs for different spoken languages X. The task is to retrieve the English text translation for a given speech query (X). The table shows the number of speech queries per spoken language X. The number of speech queries in the evaluation set varies across languages, with more queries for high-resource languages and less for low-resource languages. It is a function of the evaluation set available for different languages in the CoVoST-2 eval set. The search for the English translation is over a text database that consists of 1.6M English sentences. The text DB contains the actual English translations and the text transcriptions from the CommonVoice English dataset. We added the extra English sentences to make the translation retrieval task harder than searching over a small database of only true English translations. See Section~\ref{sec:retrieval_tasks} for more details on X$\rightarrow$EN retrieval tasks.
\begin{table*}[]
    \centering
    \caption{We perform \textbf{zero-shot} EN$\rightarrow$Y text translation retrieval on \textbf{Out-of-domain} MUST-C dataset. The search database for each EN$\rightarrow$Y retrieval task consists of approximately 200K sentences in language Y, and the query database consists of about 4K English speech utterances. The task is to retrieve the correct text translation for the English speech queries. We report the Retrieval accuracy (R@1) and the Word Error Rate between the ground-truth and retrieved text translations. We compare our retrieval pipeline \muxlsr{}-\labse{}, with \asr{}-\labse{} and the Topline retrieval model. The \muxlsr{}-\labse{} retrieval pipeline transforms speech queries to embedding vectors using our \muxlsr{} speech encoder. Then, we match the query embedding vectors with the \labse{} text embeddings of the sentences in the search DB to retrieve the translation. The \asr{}-\labse{} retrieval pipeline first uses an English language ASR to transcribe speech queries and then uses \labse{} to perform text-to-text translation retrieval. The Topline model uses the ground-truth text transcripts for the speech queries and performs text-to-text translation retrieval tasks using \labse{}.}
    \label{tab:en_y_res_ood}
    \renewcommand{\arraystretch}{1.3}
    \resizebox{.99\linewidth}{!}{%
    \begin{tabular}{lrrrrrrrrrrrrrrrr}
\toprule
  \textbf{Y}&   \textbf{DE} &   \textbf{PT} &   \textbf{FR} &   \textbf{DE} &   \textbf{RO} &   \textbf{NL} &   \textbf{IT} &   \textbf{CS} &   \textbf{VI} &   \textbf{FA} &   \textbf{TR} &   \textbf{AR} &   \textbf{RU} &   \textbf{ZH} &\textbf{Avg.}\\\midrule
  \multicolumn{16}{l}{\muxlsr{}-\labse{} Speech(EN)$\rightarrow$Text(Y) Retrieval}\\\midrule
\textbf{R@1[\%]} & 87.4 & 88.2 & 87.1 & 86.8 & 87.3 & 86.3 & 85.6 & 85.1 & 82.4 & 82.5 & 84.1 & 83.2 & 81.3 & 77.8 & 84.6\\
\textbf{WER[\%]} &  7.0 &  6.8 &  7.3 &  7.4 &  7.5 &  7.8 &  8.5 & 10.1 & 10.2 & 12.3 & 11.7 & 13.8 & 13.4 & 21.0&10.3 \\\midrule
\multicolumn{16}{l}{\asr{}-\labse{} Speech(EN)$\rightarrow$Text(Y) Retrieval}\\\midrule
\textbf{R@1[\%]} & 88.8 & 88.6 & 88.4 & 87.9 & 87.5 & 87.0 & 86.6 & 86.4 & 83.0 & 83.8 & 84.5 & 83.8 & 82.7 & 79.0 &  85.6 \\
\textbf{WER[\%]} &  6.2 &  6.5 &  6.4 &  6.7 &  7.4 &  7.2 &  7.7 &  8.9 &  9.8 & 10.3 & 11.1 & 13.2 & 12.3 & 20.6 &   9.6 \\\midrule
\multicolumn{16}{l}{Topline \labse{} Text(EN)$\rightarrow$Text(Y) Retrieval}\\\midrule
\textbf{R@1[\%]} & 96.1 & 96.0 & 96.1 & 95.9 & 95.7 & 95.3 & 95.1 & 95.1 & 92.9 & 91.4 & 92.7 & 92.4 & 92.3 & 87.6 &  93.9 \\
\textbf{WER[\%]} &  1.8 &  1.9 &  1.8 &  1.8 &  2.2 &  2.1 &  2.5 &  3.2 &  3.1 &  6.1 &  5.2 &  6.5 &  5.0 & 10.0 &   3.8 \\
\end{tabular}}
\end{table*}

\begin{table*}[]
    \caption{We present results on \textbf{Out-of-domain} MTEDx X$\rightarrow$Y text translation retrieval tasks. For a retrieval task X\_Y, the speech queries are in language X, and the search DB consists of sentences in language Y. The task is to retrieve the correct text translation for each speech query. We report the Retrieval accuracy (R@1) and the Word Error Rate between the ground-truth and retrieved text translations. We compare our retrieval pipeline \muxlsr{}-\labse{}, with \asr{}-\labse{} and the Topline retrieval model. The \muxlsr{}-\labse{} retrieval pipeline transforms speech queries to embedding vectors using our \muxlsr{} speech encoder. Then, we match the query embedding vectors with the \labse{} text embeddings of the sentences in the search DB to retrieve the translation. The \asr{}-\labse{} retrieval pipeline first uses an ASR model for language X to transcribe speech queries and then use \labse{} to perform text-to-text translation retrieval. The Topline model uses the ground-truth text transcripts for the speech queries and performs text-to-text translation retrieval tasks using \labse{}.}
    \label{tab:x_y_res}
    \centering
    \renewcommand{\arraystretch}{1.3}
    \resizebox{.99\linewidth}{!}{%
    \begin{tabular}{lrrrrrrrrrrrr}
\toprule
  \textbf{X\_Y} & \textbf{IT\_ES} &  \textbf{IT\_EN} &  \textbf{ES\_FR} &  \textbf{ES\_IT} &  \textbf{FR\_PT} &  \textbf{ES\_PT} &  \textbf{FR\_EN} &  \textbf{PT\_ES} &  \textbf{ES\_EN} &  \textbf{PT\_EN} &  \textbf{RU\_EN}&\textbf{Avg.} \\\midrule
  \textbf{Query DB}&1.8K&2K&1.8K&270&2K&1.8K&2K&2K&1.8K&2K&1.8K&- \\
  \textbf{Search DB} &1.6M&270K&220K&250K&270K&210K&1.6M&1.6M&1.6M&210K&270K&-\\\midrule
  \multicolumn{13}{l}{\muxlsr{}-\labse{} Speech(X)$\rightarrow$Text(Y) Retrieval}\\\midrule
\textbf{R@1[\%]} &   92.2 &   87.0 &   87.5 &   84.5 &   86.7 &   85.9 &   81.0 &   80.8 &   78.6 &   74.6 &   61.2 &  81.8 \\
 \textbf{WER[\%]} &   2.8 &    5.7 &    6.2 &    6.3 &    6.3 &    6.4 &    8.3 &    9.4 &    9.6 &   12.4 &   26.0 &   9.0 \\\midrule
  \multicolumn{13}{l}{\asr{}-\labse{} Speech(X)$\rightarrow$Text(Y) Retrieval}\\\midrule
\textbf{R@1[\%]} &   92.5 &   88.2 &   90.2 &   85.7 &   87.1 &   88.3 &   82.9 &   84.7 &   82.2 &   80.1 &   69.9 &  84.7 \\
 \textbf{WER[\%]} &    2.4 &    4.4 &    4.2 &    6.1 &    5.5 &    4.5 &    6.9 &    7.1 &    7.4 &    8.8 &   17.0 &   6.8 \\\midrule
 \multicolumn{13}{l}{Topline \labse{} Text(X)$\rightarrow$Text(Y) Retrieval}\\\midrule
\textbf{R@1[\%]} &   96.1 &   93.3 &   94.4 &   91.7 &   93.5 &   94.1 &   90.9 &   94.8 &   90.5 &   91.7 &   87.6 &  92.6 \\
 \textbf{WER}[\%] &    1.0 &    2.3 &    1.8 &    2.9 &    2.0 &    1.7 &    2.9 &    1.3 &    3.2 &    2.6 &    5.4 &   2.5 \\
\end{tabular}}
\end{table*}

 Interestingly, \asr{}-\labse{} is significantly worse than \muxlsr{}-\labse{} retrieval model on retrieval tasks where the speech queries are in non-European languages. For example, on ID$\rightarrow$EN, FA$\rightarrow$EN, AR$\rightarrow$EN, ZH$\rightarrow$EN, MN$\rightarrow$EN, JA$\rightarrow$EN and TA$\rightarrow$EN retrieval tasks, \muxlsr{}-\labse{} achieves a WER of 9.5\%, 10.2\%, 13.8\%, 15.2\%, 26.0\%, 44.7\% and 57.7\% respectively compared to 23.4\%, 16.8\%, 34.3\%, 36.0\%, 41.3\%, 72.9\%, 75.0\% respectively by \asr{}-\labse{}. On average \muxlsr{}-\labse{} achieves an average WER of 22.6\% compared to 33.7\% with \asr{}-\labse{} on non-European spoken languages (X)$\rightarrow$EN translation retrieval tasks. On retrieval tasks, where speech queries are in European languages, \muxlsr{}-\labse{} performs at par with \asr{}-\labse{} retrieval pipeline. For example, on RU$\rightarrow$EN, IT$\rightarrow$EN, FR$\rightarrow$EN, ES$\rightarrow$EN, DE$\rightarrow$EN, ET$\rightarrow$EN, CY$\rightarrow$EN, NL$\rightarrow$EN, CA$\rightarrow$EN, SV$\rightarrow$EN, SL$\rightarrow$EN and LV$\rightarrow$EN translation retrieval tasks, \muxlsr{}-\labse{} achieves an average WER of 13.6\% compared to 10.2\% with \asr{}-\labse{} retrieval pipeline. These results are not surprising given the fact that for European languages (high and low-resource), the ASR system is generally better than for the non-European languages. This is due to the fact that the XLSR speech encoder, which we fine-tune on downstream ASR task using language-specific transcribed data, is pre-trained on majority European language speech data.

Finally, the topline model uses the ground-truth text transcriptions corresponding to the speech queries (X) to retrieve the English text translations. This model uses only LaBSE to perform the text(X)$\rightarrow$text(EN) retrieval task. The topline achieves an average WER of 14.5\% on non-European languages X and 4.9\% on European languages, which implies that we could not quite reach the topline performance with our \muxlsr{}-\labse{} retrieval pipeline and there is room for improvement. We believe that increasing the scale of the training data and using contrastive loss for training \muxlsr{} could result in improved performance. However, a training setup with a contrastive loss would require considerable engineering effort because of the engineering complexity involved in mining negative samples across GPUs as done for training \labse{} \cite{feng2020languageagnostic}. Drawing negative samples from the same GPU device would not be sufficient because of the small per GPU batch size owing to the large speech encoder size and long speech waveforms. Hence, we leave the exploration of contrastive learning for future work.

\subsection{Results on EN$\rightarrow$Y text translation retrieval tasks} Table~\ref{tab:en_y_res} and~\ref{tab:en_y_res_ood} shows the results on EN$\rightarrow$Y speech$\rightarrow$text retrieval tasks using \muxlsr{}-\labse{}, \asr{}-\labse{} and Topline \labse{} retrieval pipelines. We retrieve the text translation in a language Y for a given speech query in English for the EN$\rightarrow$Y retrieval tasks. In the results table, first, we show the number of English speech queries and the sentences in the search database for each language, Y.

For the CoVoST-2 EN$\rightarrow$Y retrieval tasks, we have 32K English speech queries in the query DB and 320K sentences in the search DB in language Y for each EN$\rightarrow$Y retrieval task. See Section~\ref{sec:retrieval_tasks} for more details on the EN$\rightarrow$Y CoVoST-2 retrieval tasks.

Table~\ref{tab:en_y_res} shows results on CoVoST-2 EN$\rightarrow$Y retrieval tasks. We have 32K English speech queries in the query DB and 320K sentences in the search DB in language Y for each EN$\rightarrow$Y retrieval task. See Section~\ref{sec:tasks} for more details on the EN$\rightarrow$Y CoVoST-2 retrieval tasks. We observe that \muxlsr{}-\labse{} and \asr{}-\labse{} retrieval pipelines perform at par achieving a retrieval WER of 7.6\% and 7.3\% respectively, while the Topline \labse{} text(EN)$\rightarrow$text(Y) retrieval pipeline achieves an average WER of 2.1\% across the 15 retrieval tasks. There is room for improvement. In particular, for retrieving text translations in non-European languages such as ZH, MN, JA, FA, AR, and TA, for which the average WER achieved by our proposed \muxlsr{}-\labse{} retrieval pipeline is 9.7\% compared to 2.8\% with the topline \labse{} text(EN)$\rightarrow$text(Y) retrieval. For European languages, our retrieval model achieves a WER of 6.1\% compared to 1.7\% for the topline model. Our model performs better in European languages (6.1\% WER) than non-European languages (9.7\% WER).

Table~\ref{tab:en_y_res_ood} shows EN$\rightarrow$Y retrieval results on the out-of-domain MUST-C evaluation corpus. We have the same number of 4K speech utterances in the query DB and 200K sentences in the search DB for all text translation retrieval tasks. We observe that \muxlsr{}-\labse{} perform at par with \asr{}-\labse{} retrieval pipeline, achieving an average of 10.3\% WER compared to 9.6\% achieved by the \asr{}-\labse{} retrieval pipeline on the 14 EN$\rightarrow$Y retrieval tasks. Our model achieves a WER of less than 10\% for most languages except TR, AR, RU, and ZH, for which the model achieves a WER of 11.1\%, 13.2\%, 12.3\%, and 20.6\% respectively. These WERs are approximately double the WERs, achieved by the topline \labse{} text(EN)$\rightarrow$text(Y) retrieval model. However, the WERs are at a respectable less than 20\% mark.
\begin{table*}[]    \caption{We perform \textbf{zero-shot} X$\rightarrow$EN speech translation retrieval on the VoxPopuli dataset. The speech queries are in a language X, and the search database consists of speech utterances that are translations of speech queries. Unlike text translation retrieval tasks, where the search DB is much bigger than the query DB, here, the search and the query DB have the same size. During its training, \muxlsr{} did not have access to any cross-lingual speech-to-speech associations. Hence, semantic alignment among speech utterances in different languages is an emergent property of the embedding vector space learned by \muxlsr{} via our proposed multimodal learning framework. We compare \muxlsr{}'s vector space with \XLSR{}.}
    \label{tab:x_en_s2s}
    \centering
    \renewcommand{\arraystretch}{1.3}
    \resizebox{.8\linewidth}{!}{%
    \begin{tabular}{lrrrrrrrrrrrr}
\toprule
  \textbf{X} &   \textbf{ES} &   \textbf{FR} &   \textbf{PL} &   \textbf{NL} &   \textbf{DE} &   \textbf{RO} &   \textbf{HR} &   \textbf{CS} &  \textbf{Avg.} \\
\midrule
\multicolumn{13}{l}{\muxlsr{} Speech(X)$\rightarrow$Speech(EN) Retrieval}\\\midrule
\textbf{Query \& Search DB} & 36K & 50K & 19K & 11K & 60K & 16K & 8K& 11K&- \\\midrule
 \textbf{R@1[\%]} & 97.9 & 97.8 & 97.7 & 97.5 & 96.0 & 76.0 & 53.3 & 52.8 &  83.6 \\
 \textbf{R@5[\%]} & 98.5 & 98.4 & 98.4 & 98.0 & 97.1 & 80.9 & 59.5 & 58.2 &  86.1 \\\midrule
 \multicolumn{13}{l}{\XLSR{} Speech(X)$\rightarrow$Speech(EN) Retrieval}\\\midrule
 \textbf{R@1[\%]} & -&  -& - & - & 0.0 & - & - & - & 0.0 \\
\end{tabular}}
\end{table*}

\subsection{Results on X$\rightarrow$Y text translation retrieval tasks} Table~\ref{tab:x_y_res} shows results on out-of-domain MTEDx X$\rightarrow$Y text translation retrieval tasks using \muxlsr{}-\labse{}, \asr{}-\labse{} and topline \labse{} retrieval pipelines. The table shows the speech queries and text search database combination for each pair X\_Y. We observe that \muxlsr{}-\labse{} achieves an average retrieval WER of 9\% compared to 6.8\% with \asr{}-\labse{} and 2.5\% with topline \labse{} on the 11 text translation retrieval tasks. It is unsurprising that \asr{}-\labse{} retrieval pipeline performs better than the \muxlsr{}-\labse{} model. Because, the speech queries for X$\rightarrow$Y retrieval tasks are in European languages and our European language ASR models are quite good. The results reported here confirm with the observation we made for X$\rightarrow$EN CoVoST-2 translation retrieval tasks, where \muxlsr{}-\labse{} performed better than \asr{}-\labse{} for non-European languages but not for the European languages. Note that if we had an ASR model that generated text transcripts that exactly matched the ground-truth transcripts, then the performance of \asr{}-\labse{} would be same as that of the topline model.

\subsection{Results on X$\rightarrow$EN speech translation retrieval tasks}\label{sec:st}
We observe that the \muxlsr{} speech encoder learns a semantically aligned vector space across several spoken languages. The model can retrieve the correct English speech translations corresponding to speech queries in a language X with above 96\% accuracy for X$\in$\{ES, FR, PL, NL, DE\}. For X$\in$\{RO, HR, CS\}, \muxlsr{}'s speech translation retrieval performance is lagging behind other languages. This result is not surprising because \muxlsr{} did not see any transcribed data from these three languages during training. \muxlsr{} achieves an average retrieval R@1 accuracy of 83.6\% across the 8 X$\rightarrow$EN speech translation retrieval tasks. On the other hand, \XLSR{} fails on this retrieval task. To get an utterance level speech embedding from \XLSR{}, we perform temporal mean pooling of the contextual frame-wise embeddings from the last layer of the model. From the poor retrieval results, it is evident that the \XLSR{} representation space is not semantically aligned across different languages. We achieve similarly poor results with representations from different \XLSR{} layers.
\begin{table*}[]
    \caption{We present Phoneme Error Rates, PER[\%], achieved by fine-tuning \muxlsr{} and \XLSR{} on the downstream phoneme recognition task across different languages. We use one hour of labeled training data for fine-tuning and twenty minutes of development data for model selection. We evaluate the models using one hour of testing data. The test data is unseen and only used after ASR fine-tuning for model evaluation. The train, dev, and test data splits are provided by \cite{rivire2020unsupervised} and used in previous works for fine-tuning \XLSR{} for phoneme recognition \cite{conneau2020unsupervised}.}
    \label{tab:per}
    \centering
    \renewcommand{\arraystretch}{1.3}
    \resizebox{.9\linewidth}{!}{%
    \begin{tabular}{lrrrrrrrrrrr}
\toprule
  \textbf{Model} & \textbf{ES}&\textbf{FR}&\textbf{IT}&\textbf{KY}&\textbf{NL}&\textbf{RU}&\textbf{SV}&\textbf{TR}&\textbf{TT}&\textbf{Avg.}\\\midrule
  \XLSR{} \cite{conneau2020unsupervised}&\textbf{2.9} &\textbf{5.0} &5.7 &6.1 &\textbf{5.8} &8.1 &12.2 &7.1 &5.1 & 6.4\\
  \XLSR{} (Our Fine-tuning setup) &3.5&5.6&5.9&\textbf{5.7}&6.8&9.5&8.1&7.8&\textbf{4.5}&6.4\\
  \muxlsr{} (This work) &4.4&5.4&\textbf{5.4}&7.7&6.0&\textbf{6.3}&\textbf{7.7}&\textbf{6.5}&6.9 &\textbf{6.2}\\
\end{tabular}}
\end{table*}


\section{Downstream Tasks: Sequence-to-Sequence Modeling}\label{sec:per} 
\subsection{Phoneme Recognition}
Table~\ref{tab:per} shows the phoneme error rates (PER) achieved by \muxlsr{} and \XLSR{} on nine Commonvoice languages. We observe that \muxlsr{} is comparable with \XLSR{} on phoneme recognition task achieving an average PER of 6.2\% compared to 6.4\% achieved by \XLSR{} across the nine target languages, namely. See Section~\ref{sec:seq_to_seq} for details about the task and the data used for Fine-tuning \muxlsr{} and \XLSR{}.
\subsection{Automatic Speech Recognition} 
\label{sec:asr}
Table~\ref{tab:wer} shows the Word Error Rates (WER) achieved by Fine-tuning \muxlsr{} and \XLSR{} on nine languages. We observe that \muxlsr{} performs at par with \XLSR{} achieving an average WER of 24.3\% compared to 25.8\% achieved by \XLSR{}. Interestingly, on the out-of-domain Arabic (AR) language, which is drawn from the MGB2 \cite{https://doi.org/10.48550/arxiv.1609.05625} news broadcast corpus (different from the \textit{read speech} CoVo corpus used to Pre-train \muxlsr{}), \muxlsr{} performs better that \XLSR{}. 

The fact that sequence-to-sequence modeling results (ASR \& Phoneme Recognition) are at par with \XLSR{} implies that \muxlsr{} in addition to being useful for zero-shot cross-lingual text and speech translation retrieval (Section~\ref{sec:results_discuss}) can also be used for sequence generation tasks like ASR.
\begin{table*}[]
    \caption{We present Word Error Rates, WER[\%], achieved by fine-tuning \muxlsr{} and \XLSR{} on the downstream speech recognition task across different languages.}
    \label{tab:wer}
    \centering
    \renewcommand{\arraystretch}{1.3}
    \resizebox{.9\linewidth}{!}{%
    \begin{tabular}{lrrrrrrrrrrr}
\toprule
  \textbf{Model} & \textbf{ES}&\textbf{FR}&\textbf{IT}&\textbf{KY}&\textbf{NL}&\textbf{RU}&\textbf{SV}&\textbf{TR}&\textbf{TT}&AR&\textbf{Avg.}\\\midrule
  \XLSR{} (Our Fine-tuning setup)&\textbf{16.2}&31.9&\textbf{16.2}&\textbf{24.2}&19.8&28.5&27.2&26.7&21.1&\textbf{46.5}&25.8\\
  \muxlsr{} (This work)&16.4&\textbf{29.4}&18.2&31.5&\textbf{17.8}&\textbf{25.6}&\textbf{18.6}&\textbf{21.4}&30.1&\textbf{43.9}&\textbf{24.3}\\
\end{tabular}}
\end{table*}

\section{Empirical analysis of various design choices}
\label{sec:anal}
In this section, we study various design decisions that went into creating the \muxlsr{} speech encoder.

\subsection{Loss and pooling functions}  While detailing \muxlsr{} in Section~\ref{sec:samuxlsr}, we mentioned that we use the Self-Attention pooling method to construct an utterance-level speech embedding from acoustic frame-level contextual embedding vectors. Also, we use the cosine distance loss for training \muxlsr{}. Table~\ref{tab:loss_pool} shows that combining cosine distance loss and the Self-Attention pooling method is better than combining other loss functions and pooling methods. We train \muxlsr{} with L1, L2, and cosine distance losses and compare its average text translation retrieval performance across the 21 X$\rightarrow$EN CoVoST-2 retrieval tasks. Also, we compare the retrieval performance with Mean, Max, and Self-Attention pooling strategies. Three loss functions with three pooling strategies lead to nine possible training configurations. For quick analysis, we train \muxlsr{} on 8 V100-32GB GPUs for 100K iterations on a subset $\mathcal{D}_{S}$ of the complete multilingual transcribed training data $\mathcal{D}$. $\mathcal{D}_{S}$ is constructed by randomly sampling 400K training examples from $\mathcal{D}$. \muxlsr{} with Self-Attention pooling method and trained with cosine distance loss reaches an average retrieval R@1 accuracy of 48.8\%, which is better than the other 8 training configurations.
\begin{table}[]
    \centering
    \caption{Avg. retrieval Performance on 21 X$\rightarrow$EN text translation retrieval tasks for different combinations of loss and pooling functions}
    \label{tab:loss_pool}
    \renewcommand{\arraystretch}{1.3}
    \resizebox{.8\linewidth}{!}{%
\begin{tabular}{llrrrr}
\toprule
Loss & Pooling &  R@5 [\%] &  R@1 [\%] &  WER [\%] \\
\midrule
  L1 &     Max &     52.2 &     44.0   &     50.9 \\
  L1 &    Mean &     52.9 &     44.6 &     49.9 \\
  L1 &     Att. &     54.0 &     45.6 &     48.8 \\
 Cos &     Max &     55.4 &     46.6 &     47.5 \\
  L2 &     Max &     55.6 &     46.8 &     47.3 \\
 Cos &    Mean &     56.3 &     47.6 &     46.2 \\
  L2 &    Mean &     57.2 &     48.2 &     45.4 \\
  L2 &     Att. &     57.6 &     48.6 &     45.3 \\
  \textbf{Cos} &     \textbf{Att.} &     \textbf{58.0} &     \textbf{48.8}  &     \textbf{44.6} \\
\end{tabular}}
\end{table}

\subsection{Data Re-balancing Smoothing parameter $\alpha$}\label{sec:alpha} This section studies the effect on the model's average retrieval performance across 21 X$\rightarrow$EN retrieval tasks when we train the model with re-balanced training data according to Equation~\ref{eq:2}. The smoothing parameter $\alpha$ is the only hyper-parameter in the data re-balancing equation. First, we construct several re-balanced multilingual transcribed speech datasets corresponding to different values of $\alpha$. Then, we randomly sample 400K utterances from re-balanced datasets for \muxlsr{} model training. We train \muxlsr{} using cosine distance loss function for 100K iterations on 8 V100-32GB GPUs.
\begin{table}[]
    \centering
    \caption{Avg. retrieval performance on 21 X$\rightarrow$EN text translation retrieval tasks for different values of $\alpha$}
    \label{tab:alphas}
    \renewcommand{\arraystretch}{1.3}
    \resizebox{.7\linewidth}{!}{%
    \begin{tabular}{rrrrrr}
\toprule
 $\alpha$ &  R@5 [\%] &  R@1 [\%]&  WER [\%] \\
\midrule
  1.00 &     58.0 &     48.8  &    44.6 \\
  0.70 &     70.3 &     60.5  &     32.2 \\
  0.30 &     79.3 &     69.5  &     22.8 \\
  0.10 &     81.6 &     71.7  &     20.5 \\
  0.01 &     81.9 &     72.0  &     19.9 \\
  \textbf{0.05} &     \textbf{82.2} &     \textbf{72.4} &     \textbf{19.6} \\
\end{tabular}}
\end{table}

\begin{table}[]
    \centering
    \caption{Avg. retrieval performance on 7 X$\rightarrow$EN text translation retrieval tasks for different $\alpha$s. The speech queries are in low-resource languages}
    \label{tab:alphas_lr}
    \renewcommand{\arraystretch}{1.3}
    \resizebox{.7\linewidth}{!}{%
    \begin{tabular}{rrrrrr}
\toprule
 $\alpha$ &  R@5 [\%] &  R@1 [\%] &  WER [\%] \\
\midrule
  1.00 &     32.1 &     23.8  &     72.1 \\
  \textbf{0.05} &     \textbf{71.9} &     \textbf{61.4} &     \textbf{29.7} \\
\end{tabular}}
\end{table}

\begin{table}[]
    \centering
    \caption{Avg. retrieval performance on 5 X$\rightarrow$EN text translation retrieval tasks for different $\alpha$s. The speech queries are in high-resource languages}
    \label{tab:alphas_hr}
    \renewcommand{\arraystretch}{1.3}
    \resizebox{.7\linewidth}{!}{%
    \begin{tabular}{rrrrrr}
\toprule
 $\alpha$ &  R@5 [\%] &  R@1 [\%] &  WER [\%] \\
\midrule
  0.05 &     92.0 &     85.0  &      9.4 \\
  \textbf{1.00} &     \textbf{93.8} &     \textbf{87.5} &      \textbf{7.3} \\
\end{tabular}}
\end{table}

We observe in Table~\ref{tab:alphas} that the models trained with re-balanced data ($\alpha<1.0$) achieve significantly better average retrieval accuracy across the 21 X$\rightarrow$EN text translation retrieval tasks than the model trained with no re-balancing ($\alpha = 1.0$). We achieve the best performance with $\alpha = 0.05$, where the model's average retrieval accuracy R@1 is 72.4\% compared to 48.8\% achieved by \muxlsr{} trained on the original dataset without any re-balancing. The massive boost in retrieval performance is due to the model doing much better on X$\rightarrow$EN retrieval tasks where speech queries are in low-resource languages, which implies that the model was indeed under-fitting on low-resource languages due to the data imbalance in the training set of \muxlsr{}. Table~\ref{tab:alphas_lr} shows that \muxlsr{} trained with data re-balancing ($\alpha=0.05$) achieves an average retrieval R@1 accuracy of 61.4\% compared to 23.8\% achieved by \muxlsr{} trained on unbalanced training set ($\alpha=1.0$). Also, Table~\ref{tab:alphas_hr} shows that there is a negligible performance difference for different $\alpha$s on X$\rightarrow$EN tasks when speech queries are in high-resource languages.
\begin{table}[]
    \centering
    \caption{Avg. retrieval performance on 21 X$\rightarrow$EN text translation retrieval tasks for different training data}
    \label{tab:setups}
    \renewcommand{\arraystretch}{1.3}
    \resizebox{.8\linewidth}{!}{%
    \begin{tabular}{lrrrrr}
\toprule
   Model &  R@5 [\%] &  R@1 [\%] &  WER [\%] \\
\midrule
\muxlsr{}\_T2 &     49.9 &     41.3  &     54.6 \\
\muxlsr{}\_T3 &     79.7 &     69.5 &     22.7 \\
\textbf{\muxlsr{}\_T1} &     \textbf{82.2} &     \textbf{72.4} &     \textbf{19.6} \\
\end{tabular}}
\end{table}

\begin{table}[]
    \centering
    \caption{Avg. retrieval performance on 7 X$\rightarrow$EN text translation retrieval tasks for different training data. The speech queries are in low-resource languages}
    \label{tab:setups_lr}
     \renewcommand{\arraystretch}{1.3}
     \resizebox{.8\linewidth}{!}{%
    \begin{tabular}{lrrrrr}
\toprule
   Model &  R@5 [\%] &  R@1 [\%] &  WER [\%] \\
\midrule
\muxlsr{}\_T2 &     15.5 &      9.2 &     91.4 \\
\muxlsr{}\_T3 &     67.3 &     55.7  &     36.1 \\
\textbf{\muxlsr{}\_T1} &     \textbf{71.9} &     \textbf{61.4} &     \textbf{29.7} \\
\end{tabular}}
\end{table}

\begin{table}[]
    \centering
    \caption{Avg. retrieval performance on 5 X$\rightarrow$EN text translation retrieval tasks for different training data. The speech queries are in high-resource languages}
    \label{tab:setups_hr}
    \renewcommand{\arraystretch}{1.3}
    \resizebox{.8\linewidth}{!}{%
    \begin{tabular}{lrrrrr}
\toprule
   Model &  R@5 [\%] &  R@1 [\%] &  WER [\%] \\
\midrule
\muxlsr{}\_T1 &     92.0 &     85.0  &      9.4 \\
\muxlsr{}\_T2 &     91.9 &     84.9  &      9.2 \\
\textbf{\muxlsr{}\_T3} &     \textbf{92.3} &     \textbf{85.7}  &      \textbf{8.7} \\
\end{tabular}}
\end{table}

\subsection{Training Data} In Section~\ref{subsec:d}, we mention that we train \muxlsr{} with multilingual transcribed speech data collected from the CoVo dataset. In this section, we study the effect of training \muxlsr{} with paired speech-translation data. We train \muxlsr{} using three different training datasets: 1) Transcribed multilingual speech in 25 languages from the CoVo dataset, which we refer to as the training setup T1, and the model trained with this setup as \muxlsr{}\_T1, 2)  The 22 X$\rightarrow$EN CoVoST-2 \cite{wang2020covost} speech-translation training sets, where speech utterances are paired with their corresponding English text translations. We refer to that as the training setup T2, and the model trained with this setup as \muxlsr{}\_T2. 3) A combination of both T1 and T2. We refer to the model trained with this setup as \muxlsr{}\_T3. Also, we re-balance the different training datasets using $\alpha=0.05$ and then randomly pick 400K examples for training. Finally, we train the model for 100K iterations on 8 V100-32GB GPUs.

Table~\ref{tab:setups} shows average retrieval performance on 21 X$\rightarrow$EN retrieval tasks achieved by \muxlsr{} trained with the three different training setups mentioned above. We observe that \muxlsr{}\_T1 achieves the best retrieval performance out of the three models, which implies that we can train \muxlsr{} with just multilingual transcribed speech. Furthermore, table~\ref {tab:setups_lr} shows that \muxlsr{}\_T1 is notably better for X$\rightarrow$EN tasks when speech queries are in low-resource languages. For speech queries in high-resource languages, the performance difference among the three models is negligible. See Table~\ref{tab:setups_hr} for X$\rightarrow$EN retrieval tasks, when speech queries are in high-resource languages.

\section{Conclusion}
We proposed a semantically-aligned multimodal (joint speech-text) utterance-level cross-lingual speech representation (\muxlsr{}) learning framework in this work. We show that just by using multilingual transcribed speech to train the proposed representation learning model, cross-lingual alignments between speech utterances and their text and speech translations emerge in the model's learned embedding vector space. 

We show that unlike \XLSR{} (a speech-only multilingual speech encoder), \muxlsr{} in combination with language-agnostic BERT sentence encoder \labse{} can perform zero-shot speech-to-text and speech-to-speech translation retrieval across several spoken and written languages. Furthermore, we show that \muxlsr{} performs at par with \XLSR{} on sequence-to-sequence modeling tasks such as ASR and Phoneme Recognition. In the future, we will extend our multimodal learning framework for the task of zero-shot speech translation and large-scale speech-to-text data mining to create parallel speech-text translation datasets for training speech translation models.

\section*{Acknowledgments}
This work uses HPC resources of IDRIS under the allocation AD011012527 made by GENCI. We thank Nauman Dawalatabad and Yuan Gong from MIT CSAIL spoken language systems lab for reviewing the paper and provide helpful comments.

\bibliographystyle{IEEEtran}
\bibliography{bbl}

\begin{thebibliography}{10}
\providecommand{\url}[1]{#1}
\csname url@samestyle\endcsname
\providecommand{\newblock}{\relax}
\providecommand{\bibinfo}[2]{#2}
\providecommand{\BIBentrySTDinterwordspacing}{\spaceskip=0pt\relax}
\providecommand{\BIBentryALTinterwordstretchfactor}{4}
\providecommand{\BIBentryALTinterwordspacing}{\spaceskip=\fontdimen2\font plus
\BIBentryALTinterwordstretchfactor\fontdimen3\font minus
  \fontdimen4\font\relax}
\providecommand{\BIBforeignlanguage}[2]{{%
\expandafter\ifx\csname l@#1\endcsname\relax
\typeout{** WARNING: IEEEtran.bst: No hyphenation pattern has been}%
\typeout{** loaded for the language `#1'. Using the pattern for}%
\typeout{** the default language instead.}%
\else
\language=\csname l@#1\endcsname
\fi
#2}}
\providecommand{\BIBdecl}{\relax}
\BIBdecl

\bibitem{Baevski2020wav2vec}
A.~Baevski, H.~Zhou, A.~Mohamed, and M.~Auli, ``wav2vec 2.0: A framework for
  self-supervised learning of speech representations,'' \emph{arXiv preprint
  arXiv:abs/2006.11477}, 2020.

\bibitem{chung2020generative}
Y.-A. Chung and J.~Glass, ``Generative pre-training for speech with
  autoregressive predictive coding,'' in \emph{ICASSP 2020-2020 IEEE
  International Conference on Acoustics, Speech and Signal Processing
  (ICASSP)}.\hskip 1em plus 0.5em minus 0.4em\relax IEEE, 2020, pp. 3497--3501.

\bibitem{npc}
\BIBentryALTinterwordspacing
A.~H. Liu, Y.-A. Chung, and J.~Glass, ``Non-autoregressive predictive coding
  for learning speech representations from local dependencies,'' 2020.
  [Online]. Available: \url{https://arxiv.org/abs/2011.00406}
\BIBentrySTDinterwordspacing

\bibitem{pase}
\BIBentryALTinterwordspacing
S.~Pascual, M.~Ravanelli, J.~Serrà, A.~Bonafonte, and Y.~Bengio, ``Learning
  problem-agnostic speech representations from multiple self-supervised
  tasks,'' 2019. [Online]. Available: \url{https://arxiv.org/abs/1904.03416}
\BIBentrySTDinterwordspacing

\bibitem{schneider2019}
\BIBentryALTinterwordspacing
S.~Schneider, A.~Baevski, R.~Collobert, and M.~Auli, ``wav2vec: Unsupervised
  pre-training for speech recognition,'' 2019. [Online]. Available:
  \url{https://arxiv.org/abs/1904.05862}
\BIBentrySTDinterwordspacing

\bibitem{khurana2020convolutional}
\BIBentryALTinterwordspacing
S.~Khurana, A.~Laurent, W.-N. Hsu, J.~Chorowski, A.~Lancucki, R.~Marxer, and
  J.~Glass, ``A convolutional deep markov model for unsupervised speech
  representation learning,'' 2020. [Online]. Available:
  \url{https://arxiv.org/abs/2006.02547}
\BIBentrySTDinterwordspacing

\bibitem{conneau2020unsupervised}
\BIBentryALTinterwordspacing
A.~Conneau, A.~Baevski, R.~Collobert, A.~Mohamed, and M.~Auli, ``Unsupervised
  cross-lingual representation learning for speech recognition,'' 2020.
  [Online]. Available: \url{https://arxiv.org/abs/2006.13979}
\BIBentrySTDinterwordspacing

\bibitem{hsu2021hubert}
W.-N. Hsu, B.~Bolte, Y.-H.~H. Tsai, K.~Lakhotia, R.~Salakhutdinov, and
  A.~Mohamed, ``Hubert: Self-supervised speech representation learning by
  masked prediction of hidden units,'' \emph{IEEE/ACM Transactions on Audio,
  Speech, and Language Processing}, vol.~29, pp. 3451--3460, 2021.

\bibitem{babu2021xlsr}
\BIBentryALTinterwordspacing
A.~Babu, C.~Wang, A.~Tjandra, K.~Lakhotia, Q.~Xu, N.~Goyal, K.~Singh, P.~von
  Platen, Y.~Saraf, J.~Pino, A.~Baevski, A.~Conneau, and M.~Auli, ``Xls-r:
  Self-supervised cross-lingual speech representation learning at scale,''
  2021. [Online]. Available: \url{https://arxiv.org/abs/2111.09296}
\BIBentrySTDinterwordspacing

\bibitem{chen2021wavlm}
S.~Chen, C.~Wang, Z.~Chen, Y.~Wu, S.~Liu, Z.~Chen, J.~Li, N.~Kanda,
  T.~Yoshioka, X.~Xiao \emph{et~al.}, ``Wavlm: Large-scale self-supervised
  pre-training for full stack speech processing,'' \emph{arXiv preprint
  arXiv:2110.13900}, 2021.

\bibitem{w2v_bert}
\BIBentryALTinterwordspacing
Y.-A. Chung, Y.~Zhang, W.~Han, C.-C. Chiu, J.~Qin, R.~Pang, and Y.~Wu,
  ``W2v-bert: Combining contrastive learning and masked language modeling for
  self-supervised speech pre-training,'' 2021. [Online]. Available:
  \url{https://arxiv.org/abs/2108.06209}
\BIBentrySTDinterwordspacing

\bibitem{bapna2022mslam}
A.~Bapna, C.~Cherry, Y.~Zhang, Y.~Jia, M.~Johnson, Y.~Cheng, S.~Khanuja,
  J.~Riesa, and A.~Conneau, ``mslam: Massively multilingual joint pre-training
  for speech and text,'' \emph{arXiv preprint arXiv:2202.01374}, 2022.

\bibitem{laser_old}
\BIBentryALTinterwordspacing
H.~Schwenk and M.~Douze, ``Learning joint multilingual sentence representations
  with neural machine translation,'' 2017. [Online]. Available:
  \url{https://arxiv.org/abs/1704.04154}
\BIBentrySTDinterwordspacing

\bibitem{Artetxe_2019}
\BIBentryALTinterwordspacing
M.~Artetxe and H.~Schwenk, ``Massively multilingual sentence embeddings for
  zero-shot cross-lingual transfer and beyond,'' \emph{Transactions of the
  Association for Computational Linguistics}, vol.~7, pp. 597--610, nov 2019.
  [Online]. Available: \url{https://doi.org/10.1162%2Ftacl_a_00288}
\BIBentrySTDinterwordspacing

\bibitem{feng2020languageagnostic}
\BIBentryALTinterwordspacing
F.~Feng, Y.~Yang, D.~Cer, N.~Arivazhagan, and W.~Wang, ``Language-agnostic bert
  sentence embedding,'' 2020. [Online]. Available:
  \url{https://arxiv.org/abs/2007.01852}
\BIBentrySTDinterwordspacing

\bibitem{fair_mining}
\BIBentryALTinterwordspacing
H.~Schwenk, ``Filtering and mining parallel data in a joint multilingual
  space,'' 2018. [Online]. Available: \url{https://arxiv.org/abs/1805.09822}
\BIBentrySTDinterwordspacing

\bibitem{wiki_matrix}
\BIBentryALTinterwordspacing
H.~Schwenk, V.~Chaudhary, S.~Sun, H.~Gong, and F.~Guzmán, ``Wikimatrix: Mining
  135m parallel sentences in 1620 language pairs from wikipedia,'' 2019.
  [Online]. Available: \url{https://arxiv.org/abs/1907.05791}
\BIBentrySTDinterwordspacing

\bibitem{cc_matrix}
\BIBentryALTinterwordspacing
H.~Schwenk, G.~Wenzek, S.~Edunov, E.~Grave, and A.~Joulin, ``Ccmatrix: Mining
  billions of high-quality parallel sentences on the web,'' 2019. [Online].
  Available: \url{https://arxiv.org/abs/1911.04944}
\BIBentrySTDinterwordspacing

\bibitem{schwenk2017learning}
\BIBentryALTinterwordspacing
H.~Schwenk and M.~Douze, ``Learning joint multilingual sentence representations
  with neural machine translation,'' 2017. [Online]. Available:
  \url{https://arxiv.org/abs/1704.04154}
\BIBentrySTDinterwordspacing

\bibitem{shwenk2019}
\BIBentryALTinterwordspacing
M.~Artetxe and H.~Schwenk, ``Massively multilingual sentence embeddings for
  zero-shot cross-lingual transfer and beyond,'' \emph{Transactions of the
  Association for Computational Linguistics}, vol.~7, p. 597–610, Nov 2019.
  [Online]. Available: \url{http://dx.doi.org/10.1162/tacl_a_00288}
\BIBentrySTDinterwordspacing

\bibitem{schwenk2020ccmatrix}
\BIBentryALTinterwordspacing
H.~Schwenk, G.~Wenzek, S.~Edunov, E.~Grave, and A.~Joulin, ``Ccmatrix: Mining
  billions of high-quality parallel sentences on the web,'' 2019. [Online].
  Available: \url{https://arxiv.org/abs/1911.04944}
\BIBentrySTDinterwordspacing

\bibitem{gu-etal-2019-improved}
\BIBentryALTinterwordspacing
J.~Gu, Y.~Wang, K.~Cho, and V.~O. Li, ``Improved zero-shot neural machine
  translation via ignoring spurious correlations,'' in \emph{Proceedings of the
  57th Annual Meeting of the Association for Computational Linguistics}.\hskip
  1em plus 0.5em minus 0.4em\relax Florence, Italy: Association for
  Computational Linguistics, Jul. 2019, pp. 1258--1268. [Online]. Available:
  \url{https://aclanthology.org/P19-1121}
\BIBentrySTDinterwordspacing

\bibitem{arivazhagan2019missing}
N.~Arivazhagan, A.~Bapna, O.~Firat, R.~Aharoni, M.~Johnson, and W.~Macherey,
  ``The missing ingredient in zero-shot neural machine translation,''
  \emph{arXiv preprint arXiv:1903.07091}, 2019.

\bibitem{duquenne2021multimodal}
P.-A. Duquenne, H.~Gong, and H.~Schwenk, ``Multimodal and multilingual
  embeddings for large-scale speech mining,'' \emph{Advances in Neural
  Information Processing Systems}, vol.~34, 2021.

\bibitem{vaswani2017attention}
\BIBentryALTinterwordspacing
A.~Vaswani, N.~Shazeer, N.~Parmar, J.~Uszkoreit, L.~Jones, A.~N. Gomez,
  L.~Kaiser, and I.~Polosukhin, ``Attention is all you need,'' 2017. [Online].
  Available: \url{https://arxiv.org/abs/1706.03762}
\BIBentrySTDinterwordspacing

\bibitem{safari2020self}
P.~Safari, M.~India, and J.~Hernando, ``Self-attention encoding and pooling for
  speaker recognition,'' \emph{arXiv preprint arXiv:2008.01077}, 2020.

\bibitem{devlin2019bert}
\BIBentryALTinterwordspacing
J.~Devlin, M.-W. Chang, K.~Lee, and K.~Toutanova, ``Bert: Pre-training of deep
  bidirectional transformers for language understanding,'' 2018. [Online].
  Available: \url{https://arxiv.org/abs/1810.04805}
\BIBentrySTDinterwordspacing

\bibitem{lample2019cross}
G.~Lample and A.~Conneau, ``Cross-lingual language model pretraining,''
  \emph{arXiv preprint arXiv:1901.07291}, 2019.

\bibitem{wordpieces_shuster}
M.~Schuster and K.~Nakajima, ``Japanese and korean voice search,'' in
  \emph{2012 IEEE International Conference on Acoustics, Speech and Signal
  Processing (ICASSP)}, 2012, pp. 5149--5152.

\bibitem{wordpieces}
\BIBentryALTinterwordspacing
Y.~Wu, M.~Schuster, Z.~Chen, Q.~V. Le, M.~Norouzi, W.~Macherey, M.~Krikun,
  Y.~Cao, Q.~Gao, K.~Macherey, J.~Klingner, A.~Shah, M.~Johnson, X.~Liu,
  L.~Kaiser, S.~Gouws, Y.~Kato, T.~Kudo, H.~Kazawa, K.~Stevens, G.~Kurian,
  N.~Patil, W.~Wang, C.~Young, J.~Smith, J.~Riesa, A.~Rudnick, O.~Vinyals,
  G.~Corrado, M.~Hughes, and J.~Dean, ``Google's neural machine translation
  system: Bridging the gap between human and machine translation,'' 2016.
  [Online]. Available: \url{https://arxiv.org/abs/1609.08144}
\BIBentrySTDinterwordspacing

\bibitem{hugging}
\BIBentryALTinterwordspacing
T.~Wolf, L.~Debut, V.~Sanh, J.~Chaumond, C.~Delangue, A.~Moi, P.~Cistac,
  T.~Rault, R.~Louf, M.~Funtowicz, J.~Davison, S.~Shleifer, P.~von Platen,
  C.~Ma, Y.~Jernite, J.~Plu, C.~Xu, T.~L. Scao, S.~Gugger, M.~Drame, Q.~Lhoest,
  and A.~M. Rush, ``Huggingface's transformers: State-of-the-art natural
  language processing,'' 2019. [Online]. Available:
  \url{https://arxiv.org/abs/1910.03771}
\BIBentrySTDinterwordspacing

\bibitem{xlmr}
\BIBentryALTinterwordspacing
A.~Conneau, K.~Khandelwal, N.~Goyal, V.~Chaudhary, G.~Wenzek, F.~Guzmán,
  E.~Grave, M.~Ott, L.~Zettlemoyer, and V.~Stoyanov, ``Unsupervised
  cross-lingual representation learning at scale,'' 2019. [Online]. Available:
  \url{https://arxiv.org/abs/1911.02116}
\BIBentrySTDinterwordspacing

\bibitem{rebalance_data}
\BIBentryALTinterwordspacing
G.~Lample and A.~Conneau, ``Cross-lingual language model pretraining,'' 2019.
  [Online]. Available: \url{https://arxiv.org/abs/1901.07291}
\BIBentrySTDinterwordspacing

\bibitem{rebalance_data_v2}
\BIBentryALTinterwordspacing
Y.~Liu, J.~Gu, N.~Goyal, X.~Li, S.~Edunov, M.~Ghazvininejad, M.~Lewis, and
  L.~Zettlemoyer, ``Multilingual denoising pre-training for neural machine
  translation,'' 2020. [Online]. Available:
  \url{https://arxiv.org/abs/2001.08210}
\BIBentrySTDinterwordspacing

\bibitem{park2019specaugment}
D.~S. Park, W.~Chan, Y.~Zhang, C.-C. Chiu, B.~Zoph, E.~D. Cubuk, and Q.~V. Le,
  ``{SpecAugment}: A simple data augmentation method for automatic speech
  recognition,'' \emph{arXiv preprint arXiv:1904.08779}, 2019.

\bibitem{Graves12}
A.~Graves, ``Sequence transduction with recurrent neural networks,''
  \emph{arXiv preprint arXiv:abs/1211.3711}, 2012.

\bibitem{harwath2016unsupervised}
D.~Harwath, A.~Torralba, and J.~Glass, ``Unsupervised learning of spoken
  language with visual context,'' \emph{Advances in Neural Information
  Processing Systems}, vol.~29, 2016.

\bibitem{Harwath2020Learning}
\BIBentryALTinterwordspacing
D.~Harwath, W.-N. Hsu, and J.~Glass, ``Learning hierarchical discrete
  linguistic units from visually-grounded speech,'' in \emph{International
  Conference on Learning Representations}, 2020. [Online]. Available:
  \url{https://openreview.net/forum?id=B1elCp4KwH}
\BIBentrySTDinterwordspacing

\bibitem{avlnet}
\BIBentryALTinterwordspacing
A.~Rouditchenko, A.~Boggust, D.~Harwath, B.~Chen, D.~Joshi, S.~Thomas,
  K.~Audhkhasi, H.~Kuehne, R.~Panda, R.~Feris, B.~Kingsbury, M.~Picheny,
  A.~Torralba, and J.~Glass, ``Avlnet: Learning audio-visual language
  representations from instructional videos,'' 2020. [Online]. Available:
  \url{https://arxiv.org/abs/2006.09199}
\BIBentrySTDinterwordspacing

\bibitem{enwiki:939575741}
\BIBentryALTinterwordspacing
{Wikipedia contributors}, ``Word error rate --- {Wikipedia}{,} the free
  encyclopedia,'' 2020, [Online; accessed 23-April-2022]. [Online]. Available:
  \url{https://en.wikipedia.org/w/index.php?title=Word_error_rate&oldid=939575741}
\BIBentrySTDinterwordspacing

\bibitem{wang2020covost}
\BIBentryALTinterwordspacing
C.~Wang, J.~Pino, A.~Wu, and J.~Gu, ``Covost: A diverse multilingual
  speech-to-text translation corpus,'' 2020. [Online]. Available:
  \url{https://arxiv.org/abs/2002.01320}
\BIBentrySTDinterwordspacing

\bibitem{di-gangi-etal-2019-must}
\BIBentryALTinterwordspacing
M.~A. Di~Gangi, R.~Cattoni, L.~Bentivogli, M.~Negri, and M.~Turchi,
  ``{M}u{ST}-{C}: a {M}ultilingual {S}peech {T}ranslation {C}orpus,'' in
  \emph{Proceedings of the 2019 Conference of the North {A}merican Chapter of
  the Association for Computational Linguistics: Human Language Technologies,
  Volume 1 (Long and Short Papers)}.\hskip 1em plus 0.5em minus 0.4em\relax
  Minneapolis, Minnesota: Association for Computational Linguistics, Jun. 2019,
  pp. 2012--2017. [Online]. Available: \url{https://aclanthology.org/N19-1202}
\BIBentrySTDinterwordspacing

\bibitem{salesky2021multilingual}
\BIBentryALTinterwordspacing
E.~Salesky, M.~Wiesner, J.~Bremerman, R.~Cattoni, M.~Negri, M.~Turchi, D.~W.
  Oard, and M.~Post, ``The multilingual tedx corpus for speech recognition and
  translation,'' 2021. [Online]. Available:
  \url{https://arxiv.org/abs/2102.01757}
\BIBentrySTDinterwordspacing

\bibitem{wang-etal-2021-voxpopuli}
\BIBentryALTinterwordspacing
C.~Wang, M.~Riviere, A.~Lee, A.~Wu, C.~Talnikar, D.~Haziza, M.~Williamson,
  J.~Pino, and E.~Dupoux, ``{V}ox{P}opuli: A large-scale multilingual speech
  corpus for representation learning, semi-supervised learning and
  interpretation,'' in \emph{Proceedings of the 59th Annual Meeting of the
  Association for Computational Linguistics and the 11th International Joint
  Conference on Natural Language Processing (Volume 1: Long Papers)}.\hskip 1em
  plus 0.5em minus 0.4em\relax Online: Association for Computational
  Linguistics, Aug. 2021, pp. 993--1003. [Online]. Available:
  \url{https://aclanthology.org/2021.acl-long.80}
\BIBentrySTDinterwordspacing

\bibitem{rivire2020unsupervised}
\BIBentryALTinterwordspacing
M.~Rivière, A.~Joulin, P.-E. Mazaré, and E.~Dupoux, ``Unsupervised
  pretraining transfers well across languages,'' 2020. [Online]. Available:
  \url{https://arxiv.org/abs/2002.02848}
\BIBentrySTDinterwordspacing

\bibitem{watanabe2018espnet}
S.~Watanabe, T.~Hori, S.~Karita, T.~Hayashi, J.~Nishitoba, Y.~Unno, N.~{Enrique
  Yalta Soplin}, J.~Heymann, M.~Wiesner, N.~Chen, A.~Renduchintala, and
  T.~Ochiai, ``{ESPnet}: End-to-end speech processing toolkit,'' in \emph{Proc.
  Interspeech}, Sep. 2018, pp. 2207--2211.

\bibitem{arora2022espnet}
S.~Arora, S.~Dalmia, P.~Denisov, X.~Chang, Y.~Ueda, Y.~Peng, Y.~Zhang,
  S.~Kumar, K.~Ganesan, B.~Yan \emph{et~al.}, ``Espnet-slu: Advancing spoken
  language understanding through espnet,'' in \emph{ICASSP 2022-2022 IEEE
  International Conference on Acoustics, Speech and Signal Processing
  (ICASSP)}.\hskip 1em plus 0.5em minus 0.4em\relax IEEE, 2022, pp. 7167--7171.

\bibitem{Graves2006}
A.~Graves, S.~Fern{\'a}ndez, F.~J. Gomez, and J.~Schmidhuber, ``Connectionist
  temporal classification: Labelling unsegmented sequence data with recurrent
  neural networks,'' in \emph{Proc. ICML}, Jun. 2006.

\bibitem{watanabe20212020}
S.~Watanabe, F.~Boyer, X.~Chang, P.~Guo, T.~Hayashi, Y.~Higuchi, T.~Hori, W.-C.
  Huang, H.~Inaguma, N.~Kamo \emph{et~al.}, ``The 2020 espnet update: new
  features, broadened applications, performance improvements, and future
  plans,'' in \emph{2021 IEEE Data Science and Learning Workshop (DSLW)}.\hskip
  1em plus 0.5em minus 0.4em\relax IEEE, 2021, pp. 1--6.

\bibitem{https://doi.org/10.48550/arxiv.1609.05625}
\BIBentryALTinterwordspacing
A.~Ali, P.~Bell, J.~Glass, Y.~Messaoui, H.~Mubarak, S.~Renals, and Y.~Zhang,
  ``The mgb-2 challenge: Arabic multi-dialect broadcast media recognition,''
  2016. [Online]. Available: \url{https://arxiv.org/abs/1609.05625}
\BIBentrySTDinterwordspacing

\end{thebibliography}
\newpage

 



\vfill

{\appendices
\begin{table*}[]
    \centering
    \caption{Given a speech query in language X, we search over a large English database of 1.6M sentences to retrieve the top-5 translations using our proposed \muxlsr{}-\labse{} retrieval pipeline. We randomly pick five speech queries from the CoVoST-2 eval set, two in French, and one each in German, Arabic and Spanish. For each speech query, we retrieve the top-5 English translations.}
    \label{tab:quality}
    \renewcommand{\arraystretch}{1.3}
    \resizebox{.99\linewidth}{!}{%
    \begin{tabular}{lll}\toprule
         \textbf{Speech Query} & \textbf{Query Lang.} & \textbf{Top-5 Retrieved EN Translations} \\\midrule
         La chute de la cité est difficile à expliquer. &FR&1) \textbf{The fall of the city is difficult to explain}\\
        &&2) The origin of the town name is unclear.\\
        &&3) It's not easy to describe why it happened.\\
        &&4) Further history of the village is unclear.\\
        &&5) The origin of the town is not completely clear.\\\midrule
        Elle est le chef-lieu du département de l'Okano. &FR&1) It is the seat of Okanogan County. \\
        &&2) \textbf{It is the main city of the Okano District.}\\
        &&3) It is the county seat of Macon County.\\
        &&4) It is the capital of Otwock County.\\
        &&5) Its county seat is Oconto.\\\midrule
        Die Blütezeit reicht von März und April & DE & 1) \textbf{The flowering season lasts from March}\\
        vor der Bildung der Laubblätter.&&\textbf{until April, just before foliage develops.}\\
        &&2) The flowering period extends from April through June.\\
        &&3) Flowering occurs from April through July.\\
        &&4) Its flowering season is around February to April.\\
        &&5) The blooming starts in the middle of April\\
        &&and goes almost until mid May.\\\midrule
        \<.تزداد جمالاً يوماً بعد يو> &AR&1) She’s getting worse every day. \\
        &&2) It is getting better every day.\\
        &&3) It’s getting warmer day after day.\\
        &&4) \textbf{She gets prettier every day.} \\
        &&5) It’s getting colder day after day.\\\midrule
        Fue enfermera voluntaria en la I Guerra Mundial.&ES&1) \textbf{She was a volunteer nurse on World War I.}\\
        &&2) Her mother was a nurse during World War One.\\
        &&3) During World War One he served as a paramedic.\\
        &&4) During World War One he was a medical sergeant\\
        &&5) In World War One, she was a Red Cross nurse.\\
    \end{tabular}}
\end{table*}

\begin{figure*}
    \centering
    \caption{We extract the representation sequence from a Pre-trained \muxlsr{} (our proposed model) from before the attention pooling layer. Next, we compute the cosine similarity between the adjacent feature vectors to compute a sequence of distances and use a peak finding algorithm to detect the local peaks. After tuning the peak threshold in the peak finding algorithm, we observe that the peaks correspond to the underlying word boundaries.}
    \label{fig:my_label}
    \includegraphics[width=\linewidth]{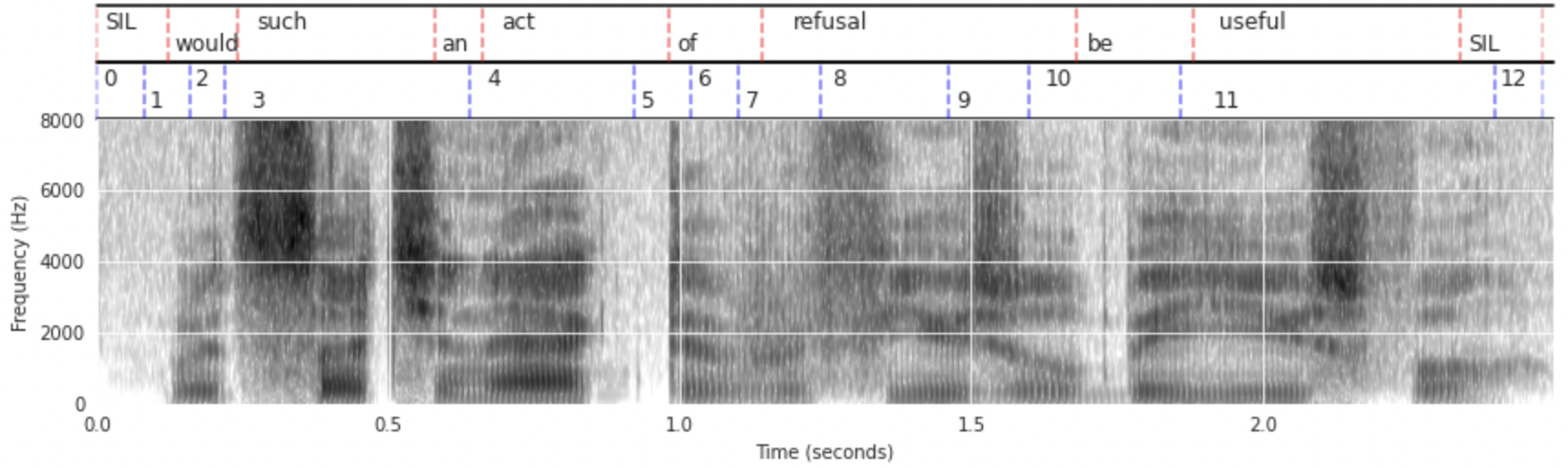}
\end{figure*}
}

\end{document}